%% file: main.tex
\documentclass[sigconf, authorversion, nonacm]{acmart}
\input{macros}
\AtBeginDocument{%
  \providecommand\BibTeX{{%
    \normalfont B\kern-0.5em{\scshape i\kern-0.25em b}\kern-0.8em\TeX}}}
    
\begin{document}

\title{Actual Causality and Responsibility Attribution in Decentralized Partially Observable Markov Decision Processes}

\author{Stelios Triantafyllou}
\affiliation{%
  \institution{Max Planck Institute for \\Software Systems}
  \country{Germany}
}
\email{strianta@mpi-sws.org}

\author{Adish Singla}
\affiliation{%
  \institution{Max Planck Institute for \\Software Systems}
  \country{Germany}
}
\email{adishs@mpi-sws.org}

\author{Goran Radanovic}
\affiliation{%
  \institution{Max Planck Institute for \\Software Systems}
  \country{Germany}
}
\email{gradanovic@mpi-sws.org}

\input{0_abstract}

\maketitle

\input{1_introduction}

\input{1.1_related_work}

\input{2_preliminaries}

\input{3_actual_causality}

\input{4_responsibility_attribution}

\input{5_experiments}

\input{6_conclusion}

\input{8_acknowledgements}

\bibliographystyle{ACM-Reference-Format}
\bibliography{main}

\clearpage
\appendix
\input{7.1_causal_graph}
\input{7.2_gumbel}
\input{7.3_demo}

\input{7.4_hp-min}
\input{7.5_add_ac_comp}

\end{document}

%% file: macros.tex
\newcommand{\agentzero}{{$ \text{Ag}0$}}
\newcommand{\agentone}{{$ \text{Ag}1$}}

\newcommand{\causalsetting}{{$ (\mathcal{C}, \vec{u})$}}
\newcommand{\witness}{$ (\vec{W}, \vec{w}', \vec{a}')$}
\newcommand{\actualcause}{$ \vec{A} = \vec{a}$}
\newcommand{\acwpair}{$ (\vec{A} = \vec{a}, (\vec{W}, \vec{w}', \vec{a}'))$}
\newcommand{\acwset}{$ H_{\mathcal{D}}$}
\newcommand{\acdef}{$ \mathcal{D}$}

\usepackage{subcaption}
\usepackage{multirow}
\usepackage[export]{adjustbox}
\usepackage{enumitem}
\usepackage{xcolor}
\usepackage{geometry} \usepackage{array, makecell}%

\usepackage{color}
\usepackage{tikz}
\usetikzlibrary{arrows}

\usetikzlibrary{shapes,decorations,arrows,calc,arrows.meta,fit,positioning}
\tikzset{
    -Latex,auto,node distance =1 cm and 1 cm,semithick,
    state/.style ={ellipse, draw, minimum width = 0.7 cm},
    point/.style = {circle, draw, inner sep=0.04cm,fill,node contents={}},
    bidirected/.style={Latex-Latex,dashed},
    el/.style = {inner sep=2pt, align=left, sloped}
}

\DeclareMathOperator*{\argmax}{arg\,max}

%% file: 0_abstract.tex
\begin{abstract}
    Actual causality and a closely related concept of responsibility attribution are central to accountable decision making. Actual causality focuses on specific outcomes and aims to identify decisions (actions) that were critical in realizing an outcome of interest. Responsibility attribution is complementary and aims to identify the extent to which decision makers (agents) are responsible for this outcome. In this paper, we study these concepts under a widely used framework for multi-agent sequential decision making under uncertainty: decentralized partially observable Markov decision processes (Dec-POMDPs). Following recent works in RL that show correspondence between POMDPs and Structural Causal Models (SCMs), we first establish a connection between Dec-POMDPs and SCMs. This connection enables us to utilize a language for describing actual causality from prior work and study existing definitions of actual causality in Dec-POMDPs. Given that some of the well-known definitions may lead to counter-intuitive actual causes, we introduce a novel definition that more explicitly accounts for causal dependencies between agents' actions. We then turn to responsibility attribution based on actual causality, where we argue that in ascribing responsibility to an agent it is important to consider both the number of actual causes in which the agent participates, as well as its ability to manipulate its own degree of responsibility. Motivated by these arguments we introduce a family of responsibility attribution methods that extends prior work, while accounting for the aforementioned considerations. Finally, through a simulation-based experiment, we compare different definitions of actual causality and responsibility attribution methods. The empirical results demonstrate the qualitative difference between the considered definitions of actual causality and their impact on attributed responsibility.
\end{abstract}

%% file: 1_introduction.tex
\section{Introduction}\label{sec.intro}

Ex-post analysis of a decision making outcome, be it perceived positive or negative, is central to accountability, which is considered to be one of the pillars of trustworthy AI \cite{ethicsEU}. Such an analysis can enable us to pinpoint decisions (hereafter actions) that caused failures and assign responsibility to decision makers (hereafter agents) involved in the decision making process. When the emphasis is put on a specific outcome and circumstances, actions that were critical in realizing this outcome constitute {\em actual causes}. The extent to which the agents' actions were critical for the outcome of interest  determines the agents' degrees of responsibility. Both actual causality and responsibility attribution have been well studied in moral philosophy, law, AI and related fields~\cite{hume2000enquiry,lewis1974causation,hitchcock2007prevention,wright1985causation,hart1985causation,halpern2016actual,datta2015program,chockler2004responsibility,coeckelbergh2020artificial,baier2021responsibility}.

A canonical approach to actual causality is based on the but-for test, which examines the counterfactual dependence of the outcome on agents' actions. It states that an action (or more generally, a set of actions) is a but-for cause of the outcome if the outcome would not have occurred had the action (resp. the set of actions) not been taken. 
It is well-known that but-for causes do not always align with human intuition---we refer the reader to \cite{halpern2016actual} for an extensive discussion.
Given this, much of the recent work on actually causality has tried to extend but-for causes in order to capture nuances of decision making scenarios where they seem to fail. 

Some of the most influential extensions are due to Halpern and Pearl \cite{halpern2001causes,halpern2005causes,halpern2015modification}, who use Structural Causal Models (SCMs)~\cite{pearl2009causality} as a framework for reasoning about actual causality. Focusing on the modified Halpern-Pearl (HP) definition \cite{halpern2015modification}, actual causes are identified through an extended but-for test, evaluated relative to some contingency. As argued by \citet{halpern2015modification}, placing appropriate restrictions on contingencies is subtle; in the modified HP definition, contingencies can only be formed from non-causal variables set to their actual values. 

While the modified HP definition generalizes but-for causality, it may still yield counter-intuitive actual causes when applied to sequential decision making
\cite{halpern2015modification}. 
An example that illustrates this is a variant of the {\em bogus prevention} scenario \cite{hitchcock2007prevention}. In this example, we have two agents, Assassin $A$ and Bodyguard $B$, whose actions influence Victim $V$. By poisoning $V$'s coffee, $A$ can cause $V$'s death, whereas $B$ can prevent $V$ from dying by putting an antidote. 
One may ask, if $A$ decides on its action after observing the action of $B$ and only poisons $V$'s coffee if $B$ has put the antidote, which actions should constitute the actual causes of $V$'s survival, in case $B$ puts the antidote and then $A$ poisons the coffee?
As argued by~\citet{halpern2015modification},
in contrast to what our intuitions would suggest, under the modified HP definition (as well as other variants of the HP definition), $B$'s action is an actual cause. Namely, putting the antidote passes the but-for test under the contingency that $A$ poisons $V$'s coffee. 
However, this is not an answer that one would expect, since $A$ had no intention of poisoning $V$ in the first place.
To correct for this, one may resort to normality considerations and extend the HP definition accordingly~\cite{halpern2015graded}. For example, when examining causality, the extended definition would exclude the ``abnormal world'' where $B$ does not add the antidote and $A$ poisons $V$'s coffee~\cite{halpern2015modification}.

However, as we show in this paper, there are sequential decision making scenarios where the HP definitions provide counter-intuitive actual causes, even under the normality considerations. These novel scenarios demonstrate that existing definitions of actual causality (i.e., the but-for definition and the HP definitions) do not fully account for conditions under which an agent decides on its actions. These conditions generally depend on the interaction history, i.e., the previous actions of the agent or the other agents. 

In this paper, we study actual causation in decentralized partially observable Markov decision processes (Dec-POMDPs)~\cite{oliehoek2016concise}, which are widely used for modeling multi-agent interactions under uncertainty. Our goal is to utilize this framework in order to derive a novel definition of actual causality
that more explicitly accounts for causal dependencies between agents' actions and their policies. 
As a down-stream task of interest, we consider responsibility attribution based on actual causality. Our contributions are as follows.

\textbf{Framework.} By relying on the recent results in reinforcement learning~\cite{buesing2018woulda,oberst2019counterfactual}, which show the correspondence between POMDPs and SCMs, we establish a connection between Dec-POMDPs and SCMs. This allows us to study existing definitions of actual causality and responsibility attribution methods in Dec-POMDPs. 

\textbf{Formal Properties.} Using sequential decision making scenarios inspired by those from the moral philosophy literature, we argue that some of the most prominent definitions of actual causality (i.e., the but-for definition and the modified HP definition) do not fully account for causal dependencies between agents' actions. The corresponding nuances are formally captured by two novel properties:
{\em Counterfactual Eligibility} and {\em Actual Cause-Witness Minimality}. 

\textbf{New Definition of Actual Causality.}
We then propose a definition of actual causality that satisfies the two novel properties. This definition utilizes additional variables, which are a part of the standard agent modeling approach in Dec-POMDPs \cite{oliehoek2016concise} that assigns to each agent an information state specifying how the agent's policy depends on the interaction history. 

\textbf{Responsibility Attribution.} We additionally study responsibility attribution based on actual causality. We introduce a family of responsibility attribution methods that extends the responsibility attribution method of \citet{chockler2004responsibility}.
These methods take into consideration the number of actual causes an agent participates in and preserve a type of {\em performance incentive} akin to the one studied by \cite{triantafyllou2021blame}---an agent cannot reduce its own degree of responsibility by increasing the number of its actions that must be changed in order to obtain a different final outcome.

\textbf{Experimental Results.} Using a simulation-based experiment, we test the qualitative properties of different definitions of actual causation and we quantify their influence on responsibility assignments. 
The experimental results show that the modified HP definition violates the two novel properties rather frequently in one of the standard benchmarks for multi-agent RL---the card game {\em Goofspiel}. For example, for a game configuration in which agents can take 12 actions in total, we find that in the majority of trajectories, 2 or more actions (i.e., more than 16\% of actions) do not conform to {\em Counterfactual Eligibility}. Similarly, 4 or more actual causes do not conform to {\em Actual Cause-Witness Minimality}. Note that the majority of trajectories have at least 13 actual causes.
The but-for definition, which satisfies {\em Actual Cause-Witness Minimality}, violates {\em Counterfactual Eligibility} even more often than the HP definition: in the game configuration from above, 3 or more actions (i.e., 25\%) do not conform to {\em Counterfactual Eligibility}.
The results also show that these property violations can have a significant effect on agents' degrees of responsibility.
When we correct for them, the agents' degrees of responsibility change in total by up to 50\%-112\%, depending on the responsibility attribution method.

We believe that these results shed a new light on actual causality and responsibility attribution, as they showcase additional challenges related to multi-agent sequential decision making. To the best of our knowledge, this is the first work that aims to tackle these challenges.

%% file: 1.1_related_work.tex
\subsection{Related Work}

In this subsection we provide a brief overview of the most relevant prior work, categorized in three different research topics: {\em actual causality}, {\em responsibility and blame attribution}, and {\em other works}. 

{\bf Actual Causality.} Arguably the closest to this paper is a recent line of work on actual causality in AI due to Halpern and Pearl \cite{halpern2001causes,halpern2005causes,halpern2015modification}, 
who introduced different versions of the HP definition of actual causality.
Works that are closely related to the HP definitions are extensively surveyed in \cite{halpern2015modification,halpern2016actual}, and they include:  \citet{pearl1998definition}, who introduced the notion of {\em causal beam} that inspired the HP definitions;
\citet{hitchcock2001intransitivity}, who identifies a variable as an actual cause by searching for a causal path in which the variable passes the but-for test; \citet{hall2007structural}, who considers the {\em H-account} definition, which identifies a subset of actual causes identified by the HP definitions; and \citet{halpern2015graded} who extend the HP definitions by incorporating normality considerations. As we already mentioned, we extend this line of work by studying  actual causality in Dec-POMDPs, which enables us to more explicitly model causal dependencies between agents' actions. This paper is also closely related to a more recent work by \citet{baier2021game}, who model multi-agent interaction via extensive form games, accounting for the conditions under which an agent decides on its actions through information states. However, \citet{baier2021game} study orthogonal aspects, primarily focusing on responsibility attribution. In contrast, we contribute to the literature on actual causality by proposing a new definition that tackles challenges related to multi-agent sequential decision making, identified in this paper.

{\bf Responsibility and Blame Attribution.} This paper is also related to the literature on responsibility attribution in multi-agent decision making. We already mentioned
\citet{chockler2004responsibility}, who consider a causality-based notion of responsibility, and \citet{baier2021responsibility}, who provide a game-theoretic account of the forward and backward notions of responsibility from \cite{poel2011relation}.
\citet{alechina2020causality} extend the decision-oriented notion of responsibility from \citet{chockler2004responsibility} to assign responsibility to agents for the failure of a team plan \cite{micalizio2004line, witteveen2005diagnosis}. In our work, we use their method as a baseline for responsibility attribution.
\citet{yazdanpanah2019strategic} study a notion of responsibly akin to the notion of blame from \cite{chockler2004responsibility},\footnote{\citet{chockler2004responsibility} differentiate responsibility and blame. For example, one of the key difference is that an agent's degree of blame depends on its epistemic state (i.e., the agent's belief about the underlying causal model). }\label{note.degree_of_blame} and similar to this paper, they explicitly incorporate time. However, their framework is based on alternating-time temporal logic (ATL), whereas we utilize Dec-POMDPs, which are more suitable for decision making under uncertainty. 
 \citet{halpern2018towards} formalize the notions of blameworthiness and intent using actual causality; similar to the degree of blame from \cite{chockler2004responsibility} (see Footnote \ref{note.degree_of_blame}), these notions depend on the epistemic state of an agent. \citet{friedenberg2019blameworthiness} extend the notion of blameworthiness to cooperative multi-agent settings. In contrast, this paper takes the notion of responsibility defined by \citet{chockler2004responsibility} as its starting point. The work by \citet{triantafyllou2021blame} is perhaps the closest in spirit to this paper as it studies blame attribution in multi-agent Markov decision processes. However, the focus of that work is on average performance as an outcome of interest, whereas we focus on specific events along a decision making trajectory. Naturally, this paper broadly relates to (cooperative) game theory and cost sharing games~\cite{von2007theory,jain2007cost}, since attribution methods such as Shapley value~\cite{shapley201617, shapley1954method} or Banzhaf index~\cite{banzhaf1964weighted, banzhaf1968one} are often utilized for defining degrees of blame, responsibility and blameworthiness~\cite{friedenberg2019blameworthiness,triantafyllou2021blame,baier2021responsibility}. 

{\bf Other Works.} From a technical point of view, this paper closely relates to RL approaches that utilize SCMs. \citet{buesing2018woulda} leverages SCMs for policy evaluation, which in turn can improve policy search methods in model-based RL. \citet{oberst2019counterfactual} extend the framework of \citet{buesing2018woulda}, allowing for off-policy evaluation in POMDPs with stochastic transition dynamics. \citet{madumal2020explainable} utilize causal models to generate explanations for actions taken by a RL agent. \citet{tsirtsis2021counterfactual} consider a causal model of the environment based on Markov decision processes (MDPs). They use this model to find an alternative sequence of actions that maximizes the counterfactual outcome, but is within a certain Hamming distance from the original action sequence. This alternative sequence serves as a counterfactual explanation.  
We contribute to this line of work by establishing a connection between Dec-POMDPs and SCMs and utilizing it for actual causality and responsibility attribution in multi-agent sequential decision making. 
Finally, this paper relates to the recent work on counterfactual {\em credit assignment} in RL \cite{harutyunyan2019hindsight, mesnard2021counterfactual}, where the goal is to improve an agent's learning efficiency by properly crediting an action for its effect on the obtained rewards. Our focus is not on improving the learning process of an agent, but on accountability considerations. 

%% file: 2_preliminaries.tex
\section{Formal Setting and Preliminaries}\label{sec.setting}

In this section, we describe our formal setting, based on decentralized partially observable Markov decision processes (Dec-POMDPs) \cite{bernstein2002complexity, oliehoek2016concise} and structural causal models (SCMs) \cite{pearl2009causality, peters2017elements}.
We also review and adopt to our setting a language for reasoning about actual causality \cite{halpern2016actual}, and we formally model the actual causality problem in the context of multi-agent sequential decision making.

\subsection{Decentralized Partially Observable Markov Decision Processes (Dec-POMDPs)}\label{sec.decpomdp}

We consider a Dec-POMDP $\mathcal{M} = (\mathcal{S}, \{1, ..., n\}, \mathcal{A}, P, \mathcal{O},  \Omega, T, \sigma)$ with $n$ agents, where: $\mathcal{S}$ is the state space; $\{1, ..., n\}$ is the agents' set; $\mathcal{A} = \times_{i = 1}^n \mathcal{A}_i$ is the joint action space, with $\mathcal{A}_i$ being the action space of agent $i$; $P$ specifies transitions with $P(s, a, s')$ denoting the probability of the process transitioning to $s'$ from $s$ when agents $\{1,..., n\}$ take joint action $a = (a_1,..., a_n)$; $\mathcal{O} = \times_{i = 1}^n \mathcal{O}_i$ is the joint observation space, with $\mathcal{O}_i$ being the observation space of agent $i$; $\Omega$ is an observation probability function with $\Omega(s, o)$ denoting the probability of agents $\{1,..., n\}$ receiving joint observation $o = (o_1,..., o_n)$ when in state $s$; $T$ is the finite time horizon; $\sigma$ is the initial state distribution. We assume $\mathcal{S}$, $\mathcal{A}$ and $\mathcal{O}$ to be finite and discrete. For ease of notation, we additionally assume that the agents' immediate rewards are part of their observations.
Throughout the paper, we denote random variables with capital letters, e.g., $S$, $A$ and $O$.

We also consider for each agent $i$ a model $m_i = (\mathcal{I}_i, \pi_i, Z_i, Z_{i,0})$ \cite{oliehoek2016concise}, where: $\mathcal{I}_i$ is the (finite and discrete) information state space of $i$; $\pi_i$ is the policy of agent $i$, i.e., a mapping $\pi_i: \mathcal{I}_i \rightarrow \Delta(\mathcal{A}_i)$, where $\Delta(\mathcal{A}_i)$ is a probability simplex over $\mathcal{A}_i$; 
$Z_i$ is agent $i$'s information probability function with $Z_i(\imath_i, a_i, o_i, \imath'_i)$ denoting the probability of $i$'s information state changing from $\imath_i$ to $\imath'_i$, after $i$ takes action $a_i$ and observes $o_i$; $Z_{i,0}$ is $i$'s initial information probability function depending only on $o_{i,0}$.
We use $\pi_i(a_{i}|\imath_i)$ to denote the probability of agent $i$ taking action $a_i$ given information state  $\imath_i$.
The agents' joint policy is denoted by $\pi$, and we assume that $\pi(a|\imath_1, ..., \imath_n) = \pi_1(a_{1}|\imath_1) \cdot\cdot\cdot \pi_n(a_{n}|\imath_n)$.
Note that information states are a way to encode the information that an agent uses in its decision making.

\subsection{Dec-POMDPs and Structural Causal Models}\label{sec.decpomdp_scm}

Although Dec-POMDPs are a very general and useful modelling tool for multi-agent sequential decision making, they are not sufficient to reason counterfactually about alternate outcomes \cite{lewis2013counterfactuals}, and hence actual causality.\footnote{By counterfactual reasoning, we mean predicting what would have happened in a specific instance of the decision process (trajectory) had some action(s) been different.}
For instance, given a trajectory $\tau = \{(s_t, a_{1, t}, ..., a_{n, t})\}_{t=0}^{T-1}$ generated by Dec-POMDP $\mathcal{M}$ under joint policy $\pi$, we would like to predict what would have happened, had agent $i$ taken action $a'_i$ instead of action $a_{i,t}$. However, even though we have access through $\mathcal{M}$ to the probability distribution of the next state $P(s_t, (a_{1, t}, ..., a'_i ..., a_{n, t}), \cdot)$, we do not have a way to infer what would be the value of $S_{t+1}$.\footnote{Or in general the value of anything that comes (chronologically) after time-step $t$.}
Following \citet{buesing2018woulda}, to overcome this limitation we view $\mathcal{M}$ under joint policy $\pi$ as a structural causal model (SCM) $\mathcal{C}$. To do this, we express $P$, $\Omega$, $\{Z_i\}_{i\in\{1,...,n\}}$ and $\{\pi_i\}_{i \in \{1, .., n\}}$  as deterministic functions $g$ with independent noise variables $U$, such as
\begin{align}\label{eq.struct_eq}
    &S_t = g_{S_t}(S_{t-1}, A_{t-1}, U_{S_t}), 
    \quad O_t = g_{O_t}(S_{t}, U_{O_t}),\nonumber\\
    &I_{i,t} = g_{I_{i, t}}(I_{i,t-1}, A_{i,t-1}, O_{i,t}, U_{I_{i,t}}), 
    \quad A_{i,t} = g_{A_{i, t}}(I_{i,t}, U_{A_{i,t}}),
\end{align}
where $U_{S_t}$, $U_{O_t}$, $U_{I_{i,t}}$ and $U_{A_{i,t}}$ are $|\mathcal{S}|$-, $|\mathcal{O}|$-, $|\mathcal{I}_i|$- and $|\mathcal{A}_i|$- dimensional, respectively.
It can be shown that such a parameterization is always possible.\footnote{In \cite{buesing2018woulda}, they show how to represent an episodic POMDP as an SCM, and prove that this is always possible. Their results can be directly extended to Dec-POMDPs.}
Henceforth, we will refer to SCMs that are defined in this way as Dec-POMDP SCMs.
Consistent with the SCMs' terminology \cite{pearl2009causality}, we also say that state variables $S_t$, observation variables $O_t$, information variables $I_{i,t}$ and action variables $A_{i,t}$ constitute the endogenous variables of $\mathcal{C}$, $U$ are the exogenous variables, and equations \eqref{eq.struct_eq} are the model's structural equations. 
The causal graph of the Dec-POMDP SCM can be found in Appendix \ref{sec.dag}.
We can generate a trajectory $\tau = \{(s_t, a_{1, t}, ..., a_{n, t})\}_{t=0}^{T-1}$ using Dec-POMDP SCM by simply specifying a setting $\vec{u}$ for the exogenous variables in $U$, also called context, and then solving the structural equations of $\mathcal{C}$, i.e., Eq. \eqref{eq.struct_eq}.
Note that for each Dec-POMDP SCM-context pair $(\mathcal{C}, \vec{u})$, also called causal setting, there is a unique trajectory $\tau$ that can be generated in that way. Importantly, we can also find out what would have happened in $\tau$, had agent $i$ taken action $a'_i$ instead of $a_{i,t}$ in the following way:

\begin{enumerate}
    \item We perform the intervention\footnote{Interventions are also often modeled through the \textit{do}-operator \cite{pearl1995causal}, \textit{do}$(A_{i,t} = a'_i)$.} $A_{i,t} \leftarrow a'_i$ on $\mathcal{C}$, that is we replace $g_{A_{i, t}}(I_{i,t}, U_{A_i,t})$ in Eq. \eqref{eq.struct_eq} with the value $a'_i$. The resulting SCM is denoted by $\mathcal{C}^{A_{i,t} \leftarrow a'_i}$.
    \item We generate the counterfactual trajectory $\tau^{\text{cf}}$ from the causal setting $(\mathcal{C}^{A_{i,t} \leftarrow a'_i}, \vec{u})$, where $\vec{u}$ is the same context that we used to generate $\tau$.\footnote{Definitions for actual causality and responsibility, which are the focus of this paper, are relative to a causal setting \cite{chockler2004responsibility}. Therefore, we assume $\mathcal{C}$ and $\vec{u}$ to be fully known.}
\end{enumerate}

Note that when Dec-POMDP $\mathcal{M}$ or joint policy $\pi$ are stochastic, the counterfactual trajectory $\tau^{\text{cf}}$ may not be identifiable without further assumptions \cite{oberst2019counterfactual}. This is because, there may be multiple parameterizations of a Dec-POMDP SCM, i.e., multiple functions $g$ and distributions over the exogenous variables $U$, which are all able to correctly represent $\mathcal{M}$ under $\pi$,\footnote{For every state, observation, information or action variable $V$, it holds that variable $V'$ equals $V$ in distribution, where $V' = g_{V}(pa_{V}, U_{V})$, and $pa_V$ are the parents of $V$ in the causal graph.} but which suggest different counterfactual outcomes, e.g., $\tau^{\text{cf}}$. Consequentially, the choice of model can have a significant impact on claims of causality. In our experiments, we choose to focus on a particular class of SCMs, the Gumbel-Max SCMs \cite{oberst2019counterfactual}. This class of SCMs has been shown to satisfy the desirable property of counterfactual stability, which excludes a specific type of non-intuitive counterfactual outcomes. Appendix \ref{sec.gumbel} provides more details on Gumbel-Max SCMs and the counterfactual stability property.

\subsection{Actual Causality}\label{sec.ac_setting}

We now review and adopt in our formal setting a language introduced by prior work on actual causality with SCMs \cite{halpern2016actual}.
Consider a Dec-POMDP SCM $\mathcal{C}$ and a context $\vec{u}$, and the (unique) trajectory generated by the causal setting $(\mathcal{C}, \vec{u})$, $\tau = \{(s_t, a_t)\}_{t=0}^{T-1}$. We call primitive event, any formula of the form $V = v$, where $V$ is an endogenous variable in $\mathcal{C}$, i.e., state, observation, information or action variable, and $v$ is a valid value for $V$. Let $\phi$ be an event, that is any Boolean combination of primitive events. We use $(\mathcal{C}, \vec{u}) \models \phi$ to denote that $\phi$ is true in the causal setting $(\mathcal{C}, \vec{u})$, i.e., $\phi$ takes place in $\tau$.
Furthermore, given a set of interventions on action variables $\vec{A} \leftarrow \vec{a}'$, we write $(\mathcal{C}, \vec{u}) \models [\vec{A} \leftarrow \vec{a}']\phi$, if $(\mathcal{C}^{\vec{A} \leftarrow \vec{a}'}, \vec{u}) \models \phi$.
For instance, consider the counterfactual scenario in which agent $i$ takes the action $a_i'$ instead of $a_{i, t}$ in $\tau$. 
If under this counterfactual scenario the process transitions to state $s$ at time-step $t+1$, then the following statement holds
\begin{align*}
    (\mathcal{C}, \vec{u}) \models [A_{i,t} \leftarrow a_i'](S_{t+1} = s).
\end{align*}
\textbf{Actual Causality in Multi-Agent Sequential Decision Making.}
Our goal is to pinpoint the actions which caused a particular event to happen. Given a causal setting $(\mathcal{C}, \vec{u})$ and the event of interest $\phi$, we want to determine the actual causes of $\phi$ in $(\mathcal{C}, \vec{u})$. In this paper, what can be an actual cause is a conjunction of primitive events consisted only of action variables, abbreviated here as \actualcause.
We say that every conjunct of actual cause \actualcause~is part of that cause.
Furthermore, in some cases we want to define an actual cause \actualcause~w.r.t. some contingency, that is \actualcause~is an actual cause only if that contingency holds. What can be a contingency in this paper is again a conjunction of primitive events consisted only of action variables, abbreviated as $\vec{W} = \vec{w}'$. Finally, for \actualcause~to be an actual cause of $\phi$ in \causalsetting~under contingency $\vec{W} = \vec{w}'$, there has to exist a setting $\vec{a}'$, such that $(\mathcal{C}, \vec{u}) \models [\vec{A} \leftarrow \vec{a}', \vec{W} \leftarrow \vec{w}']\neg\phi$. We will often refer to $\vec{a}'$ as (counterfactual) setting. Consistent with the actual causality literature \cite{halpern2016actual}, we call the tuple \witness~a witness to the fact that \actualcause~is an actual cause of $\phi$ in \causalsetting. 

Coming back to the introduction example, \causalsetting~models the considered trajectory: $B$ puts the antidote ($B = \textit{antidote}$); $A$ poisons $V$'s coffee ($A = \textit{poison}$); $V$ \textit{survives}.
The outcome of interest $\phi$ is that $V$ \textit{survives}.
According to the HP definition, the action $B = \textit{antidote}$ consists the actual cause of $\phi$ in \causalsetting~under the contingency that $A = \textit{poison}$. Indeed, there is a counterfactual setting for $B$ such that
\begin{align*}
    (\mathcal{C}, \vec{u}) \models [B = \textit{not antidote}, A = \textit{poison}]\neg (V \textit{survives}).
\end{align*}
In Section \ref{sec.ac_definitions}, we consider several definitions of actual causality w.r.t. a causal setting $(\mathcal{C}, \vec{u})$, where $\mathcal{C}$ is always a Dec-POMDP SCM. 
More specifically, in this paper an actual causality definition \acdef~has to formally describe a process that receives as input a causal setting \causalsetting~and an event of interest $\phi$, and outputs a set of actual cause-witness pairs, i.e., a set of elements of the form \acwpair. We use $H_{\mathcal{D}}((\mathcal{C}, \vec{u}), \phi)$ (or just \acwset, when \causalsetting~and $\phi$ are implied) to denote the set of all actual cause-witness pairs of $\phi$ in \causalsetting, according to \acdef. 
We also refer to $(\mathcal{C}, \vec{u})$ as the actual world or situation. Similarly, we refer to a causal setting as the counterfactual world or scenario when it results from $(\mathcal{C}, \vec{u})$ after an intervention is performed on a subset of its action variables, e.g., $(\mathcal{C}^{\vec{A} \leftarrow \vec{a}'}, \vec{u})$.

%% file: 3_actual_causality.tex
\section{Definitions for Actual Cause}\label{sec.ac_definitions}

In this section, we analyze two of the most popular definitions of actual causality that involve counterfactuals, the ``but-for'' definition\footnote{Also known as cause-in-fact and \textit{sine qua non}.} (BF definition from now on) \cite{hart1985causation}, and the Halpern and Pearl definition (HP definition from now on) \cite{halpern2015modification}. 
We provide two counterexamples (both are new variants of the ``bogus prevention'' scenario \cite{hiddleston2005causal}) which expose several weaknesses of the two definitions.
We formally capture the insights we gain from these examples with two novel properties. Subsection \ref{sec.def_new} introduces a new definition for actual cause, which satisfies these two properties.

\subsection{The BF Definition}\label{sec.bf_def}

But-for cause is one of the fundamental definitions of causation in law \cite{hart1985causation}, and it states that \textit{$C$ is a cause of $E$ if but for $C$, $E$ would not have occurred}.
In other words, $C$ was necessary for $E$ to happen. In our setting, we formally define but-for cause as follows.

\begin{definition}\label{def.bfc}
(But-For Cause)
\actualcause~is a but-for cause of the event $\phi$ in \causalsetting~if the following conditions hold:
\begin{enumerate}[label=BFC\arabic*.]
    \item $(\mathcal{C}, \vec{u}) \models (\vec{A} = \vec{a})$ and $(\mathcal{C}, \vec{u}) \models \phi$
    \item There is a setting $\vec{a}'$ of the variables in $\vec{A}$, such that
    \begin{align*}
        (\mathcal{C}, \vec{u}) \models [\vec{A} \leftarrow \vec{a}']\neg\phi
    \end{align*}
    \item $\vec{A}$ is minimal; there is no subset $\vec{A}'$ of $\vec{A}$, such that $\vec{A}' = \vec{a}'$ satisfies \textit{BFC1} and \textit{BFC2}, where $\vec{a}'$ is the restriction of $\vec{a}$ to the variables of $\vec{A}$
\end{enumerate}
We say that $(\emptyset, \emptyset, \vec{a}')$ is a witness of \actualcause~being a but-for cause of $\phi$ in \causalsetting.
\end{definition}

\textit{BFC1} requires that both \actualcause~and $\phi$ happened in the actual world, \causalsetting. \textit{BFC2} implies the but-for condition, i.e., but for \actualcause, $\phi$ would not have occurred.
\textit{BFC3} is a minimality condition, which ensures that an actual cause does not include any non-essential elements.
Unfortunately, but-for cause does not suffice for a good definition of actual cause in the context of sequential decision making, and the next example illustrates some of the reasons.

\begin{example}\label{ex.single_bogus}
Victim $V$ dines at time-step $t$. Assassin $A$ has access to $V$'s table at time-steps $t-2$ and $t-1$, when they can choose whether to \textit{poison} or \textit{not poison} $V$'s water. $A$'s policy is to always \textit{poison} $V$'s water, unless it is already poisoned. We consider the trajectory, in which $A$ chooses to \textit{poison} $V$'s water only at time-step $t-2$, and $V$ dies from the poison at time $t$.
\end{example}

To identify a but-for cause of $V$ dying at time-step $t$, consider an intervention that sets $A$'s action at time-step $t-2$ to \textit{not poison}. If this is the only intervention, $A$ follows its policy at $t-1$ and takes action \textit{poison}, which results in the same outcome. To change the outcome, we also need to intervene at time-step $t-1$ and set $A$'s action to \textit{not poison}. This implies that action \textit{poison} taken at $t-2$ and action \textit{not poison} taken at $t-1$ form a but-for cause of $V$ dying at time $t$. We find this counter-intuitive because the action that has to be changed at time-step $t-1$ is not the one that was taken in the actual situation but the one that would have been taken in the counterfactual scenario where $A$ does \textit{not poison} the water at $t-2$. Since the conditions that influence $A$'s decision at $t-1$ change once we intervene at $t-2$, we argue that the action taken at $t-1$ should not be a part of an actual cause, but should be treated as a contingency. The following property formalizes this insight.

\begin{property}[Counterfactual Eligibility]\label{prop.ce}
We say that a definition for actual cause \acdef~satisfies \textit{Counterfactual Eligibility} if for every $(\mathcal{C}, \vec{u})$, $\phi$, \acwpair~and $A_{i,t} = a_{i,t}$, where \acwpair~is an actual cause-witness pair of $\phi$ in \causalsetting~according to \acdef, and $A_{i,t} = a_{i,t}$ is part of \actualcause, it holds that $(\mathcal{C}, \vec{u}) \models [\vec{A} \leftarrow \vec{a}', \vec{W} \leftarrow \vec{w}'] (I_{i,t} = \imath_{i,t})$, where $\imath_{i,t}$ is agent $i$'s information state in \causalsetting, i.e., $(\mathcal{C}, \vec{u}) \models (I_{i,t} = \imath_{i,t})$.
\end{property}

Property \ref{prop.ce} states that an agent's action is eligible for being a part of an actual cause if the information state under which the agent took that action in the actual world remains the same in the witness world, i.e., the counterfactual world which corresponds to the cause's witness.
As Example \ref{ex.single_bogus} suggests, the BF definition violates Property \ref{prop.ce}.

\subsection{The HP Definition}\label{sec.hp_def}

Arguably, one of the most influential accounts of causality is Halpern and Pearl's notion of actual causes in SCMs \cite{halpern2005causes}. There are three variants of the HP definition of actual causality \cite{halpern2001causes, halpern2005causes, halpern2015modification}. In this paper we consider and adopt in our setting the latest one \cite{halpern2015modification}.\footnote{All definitions and their relations are extensively discussed in \cite{halpern2016actual}.}

\begin{definition}\label{def.hp}
(HP) \actualcause~is an actual cause of the event $\phi$ in \causalsetting~if the following conditions hold:
\begin{enumerate}[label=HPC\arabic*.]
    \item $(\mathcal{C}, \vec{u}) \models (\vec{A} = \vec{a})$ and $(\mathcal{C}, \vec{u}) \models \phi$
    \item There is a set $\vec{W}$ of action variables and a setting $\vec{a}'$ of the variables in $\vec{A}$ such that if $(\mathcal{C}, \vec{u}) \models (\vec{W} = \vec{w}')$, then
    \begin{align*}
        (\mathcal{C}, \vec{u}) \models [\vec{A} \leftarrow \vec{a}', \vec{W} \leftarrow \vec{w}']\neg\phi
    \end{align*}
    \item $\vec{A}$ is minimal; there is no subset $\vec{A}'$ of $\vec{A}$, such that $\vec{A}' = \vec{a}'$ satisfies \textit{HPC1} and \textit{HPC2}, where $\vec{a}'$ is the restriction of $\vec{a}$ to the variables of $\vec{A}$
\end{enumerate}
We say that \witness~is a witness of \actualcause~being an actual cause of $\phi$ in \causalsetting.
\end{definition}

\textit{HPC1} and \textit{HPC3} are similar to \textit{BFC1} and \textit{BFC3}, respectively. \textit{HPC2} says that the but-for condition holds under the contingency $\vec{W} = \vec{w}'$, where the setting $\vec{w}'$ has the observed value of $\vec{W}$ in \causalsetting. Roughly speaking, this means that \actualcause~is an actual cause of $\phi$ in \causalsetting~if but for \actualcause, $\phi$ would not have happened, had the action variables in $\vec{W}$ been fixed to their actual values.
The main intuition behind \textit{HPC2}, and what differentiates this HP definition from its predecessors, is that \textit{``only what happens in the actual situation should matter''} \cite{halpern2016actual}.

Coming back to Example \ref{ex.single_bogus}, according to the HP definition, the actual cause of $V$ dying at time-step $t$ is the action of $A$ to \textit{poison} the water at time-step $t-2$, under the contingency that $A$ would \textit{not poison} the water at $t-1$. In other words, if we assume that in the counterfactual world where $A$ does \textit{not poison} the water at $t-2$, they also do \textit{not poison} the water at $t-1$, then the first action is considered an actual cause of $V$ dying. We find this answer more intuitive than the one given by the BF definition.
Despite the success of the HP definition in Example \ref{ex.single_bogus} as well as in many more examples in the moral philosophy literature \cite{halpern2015modification, halpern2016actual}, we illustrate with the next example that the HP definition is not sufficient for multi-agent sequential decision making.

\begin{example}\label{ex.multi_bogus}
Victim $V$ dines at time-step $t$. Bodyguard $B$, who suspects a poisonous attack, has access to $V$'s table at time-step $t-2$, when they can choose where to put an antidote, either into $V$'s \textit{water} or into $V$'s \textit{wine}. $B$ is right, indeed an assassin $A$ has access to $V$'s table at time-step $t-1$, when they can choose where to put the poison, again either into $V$'s \textit{water} or into $V$'s \textit{wine}. The poison is neutralized by the antidote only if they have been put into the same liquid, otherwise $V$ dies. We assume that $A$ observes where $B$ puts the antidote and that its intention is to poison $V$.
Therefore, $A$'s policy is to put the poison into the liquid that does not have the antidote.
Consider the trajectory, in which $B$ puts the antidote into the \textit{water} at time-step $t-2$, $A$ puts the poison into the \textit{wine} at time-step $t-1$, and $V$ dies at time $t$.
\end{example}

According to both BF and HP definitions, an actual cause of $V$ dying at $t$ is the action of $A$ poisoning the \textit{wine} at $t-1$. However, according to the HP definition, this is not the only actual cause of $V$'s death. The action of $B$ putting the antidote into the \textit{water} is also considered an actual cause, under the contingency that $A$ would poison the \textit{wine}, i.e., $B$'s action is an actual cause assuming that $A$'s action is fixed to its value in the \textit{``actual situation''}.
\footnote{Note that based on Theorem 2.3 from \cite{halpern2015modification} both older versions of the HP definition also consider $B$'s action as an actual cause of $V$'s death in Example \ref{ex.multi_bogus}.}
In particular, $B$ did put the antidote into the \textit{water} in the actual scenario, and since it is a single action, it is also minimal, hence \textit{HPC1} and \textit{HPC3} are satisfied. Furthermore, if we intervene on $B$'s action by changing it to \textit{wine}, and fix $A$'s action also to \textit{wine} (the realized action), then $V$ does not die, and hence \textit{HPC2} is satisfied. 

We find the latter actual cause to be counter-intuitive. Essentially $B$ had no control over the final outcome of this example, because of the full observability and policy assumed for $A$. This counter-intuitive result is due to the fact that the HP definition applies the minimality condition (\textit{HPC3}) only on the variables of the actual cause \actualcause, and does not include those of the contingency $\vec{W} = \vec{w}'$ (here $\vec{W} = \vec{w}'$ is in fact another actual cause). Motivated by Example \ref{ex.multi_bogus}, we introduce the following property.

\begin{property}[Actual Cause-Witness Minimality]\label{prop.min}
We say that a definition for actual cause \acdef~satisfies \textit{Actual Cause-Witness Minimality} if for every $(\mathcal{C}, \vec{u})$, $\phi$, \actualcause~and $\vec{W} = \vec{w}'$, where \actualcause~is an actual cause of $\phi$ in $(\mathcal{C}, \vec{u})$ under contingency $\vec{W} = \vec{w}'$ according to \acdef, there are no $\vec{A}'$, $\vec{W}'$ and $\vec{w}''$, such that $\vec{A}' \cup \vec{W}' \subset \vec{A} \cup \vec{W}$ and $\vec{A}' = \vec{a}'$ is an actual cause of $\phi$ in $(\mathcal{C}, \vec{u})$ under contingency $\vec{W}' = \vec{w}''$ according to \acdef, where $\vec{a}'$ is the restriction of $\vec{a}$ to $\vec{A}$. 
\end{property}

Roughly speaking, Property \ref{prop.min} extends \textit{HPC3} to also include $\vec{W}$, i.e., $\vec{A} \cup \vec{W}$ is minimal. As Example \ref{ex.multi_bogus} suggests, the HP definition violates Property \ref{prop.min}. Additionally, Appendix \ref{sec.demonstration_app} describes a scenario where the HP definition violates Property \ref{prop.ce}.

Example \ref{ex.multi_bogus} also sets the ground for arguing about the interpretation of the {\em actuality} test: \textit{``only what happens in the actual situation should matter''}. For instance, one can argue that  $A$'s action of putting the poison into the \textit{wine} is a valid contingency for the HP definition since this action did realize in the actual situation. Arguably, this interpretation is adopted in \cite{halpern2015modification}.
In contrast, we argue for an interpretation that focuses not just on agents' actions, but also on their information states, i.e., conditions under which agents reach their decisions. 
Under this interpretation, $A$'s action of putting the poison into the \textit{wine} does not pass the actuality test
for the counterfactual world in which $B$ puts the antidote into the \textit{wine}. Namely, in the actual situation, $A$ put the poison into the \textit{wine} only because $B$ had put the antidote into the \textit{water}. Now, this interpretation may be restrictive if the actuality test is applied on contingencies, as it is the case with the HP definition. 
For instance, if condition \textit{HPC2} is modified accordingly, the HP definition would identify no actual causes in Example \ref{ex.single_bogus}.
However, we believe that the actuality test is not important for contingencies, but only for actual causes. This is formalized by Property \ref{prop.ce}. 
Note also that the BF definition satisfies Property \ref{prop.min} because of \textit{BFC3}.

\textbf{Normality and Defaults.} 
The notions of normality and defaults have been shown to deal with a number of examples where the HP definitions provide counter-intuitive actual causes \cite{halpern2008defaults, halpern2015graded}. However, Example \ref{ex.multi_bogus} is not one of them. More specifically, in this example there are two possible worlds, one where $B$ and $A$ put the antidote and the poison into the same liquid, and one where they put them into different liquids. 
Given the intentions of $A$ and that $A$ observes $B$'s action in this example, one may consider the former world less normal than the latter one.
All $3$ HP definitions, when extended to account for the aforementioned normality ordering, they provide no actual causes, although there is arguably one. We conclude, that despite the usefulness of these notions, they do not address the shortcomings of the core definition described above.

\subsection{A New Definition for Actual Cause}\label{sec.def_new}

We extend the BF definition with the notion of contingencies and implement the insights we gain from Examples \ref{ex.single_bogus} and \ref{ex.multi_bogus}, to propose a new definition for actual cause. Intuitively, our definition takes a but-for cause and splits its set of (action) variables into two subsets: the actual cause and the contingency. The partition is based on whether the conditions under which these actions were taken, change between the actual world and the witness one.

\begin{definition}\label{def.ours}
(Actual Cause)
\actualcause~is an actual cause of the event $\phi$ in \causalsetting, under the contingency $\vec{W} = \vec{w}'$ if the following conditions hold:
\begin{enumerate}[label=AC\arabic*.]
    \item There is a setting $\vec{a}'$ of the variables in $\vec{A}$, such that $\vec{A} = \vec{a} \wedge \vec{W} = \vec{w}$ is a but-for cause of $\phi$ in \causalsetting, and also satisfies condition \textit{BFC2} with setting $(\vec{a}', \vec{w}')$
    \item For every agent $i$ and time-step $t$ such that $A_{i,t} \in \vec{A}$ and $(\mathcal{C}, \vec{u}) \models (I_{i,t} = \imath_{i,t})$, it holds that
    \begin{align*}
        (\mathcal{C}, \vec{u}) \models [\vec{A} \leftarrow \vec{a}', \vec{W} \leftarrow \vec{w}'] (I_{i,t} = \imath_{i, t})
    \end{align*}
    \item For every agent $i$ and time-step $t$ such that $A_{i,t} \in \vec{W}$ and $(\mathcal{C}, \vec{u}) \models (I_{i,t} = \imath_{i,t})$, it holds that
    \begin{align*}
        (\mathcal{C}, \vec{u}) \models [\vec{A} \leftarrow \vec{a}', \vec{W} \leftarrow \vec{w}'] \neg (I_{i,t} = \imath_{i,t})
    \end{align*}
\end{enumerate}
We say that \witness~is a witness of \actualcause~being an actual cause of $\phi$ in \causalsetting.
\end{definition}

\textit{AC1} says that combined $\vec{A} = \vec{a}$ and $\vec{W} = \vec{w}$ should form a but-for cause of $\phi$ in \causalsetting~using the settings $\vec{a}'$ and $\vec{w}'$. According to the BF definition, this means that the following conditions should hold: 
\begin{enumerate}
    \item $(\mathcal{C}, \vec{u}) \models (\vec{A} = \vec{a})$, $(\mathcal{C}, \vec{u}) \models (\vec{W} = \vec{w})$ and $(\mathcal{C}, \vec{u}) \models \phi$
    \item $(\mathcal{C}, \vec{u}) \models [\vec{A} \leftarrow \vec{a}', \vec{W} \leftarrow \vec{w}']\neg\phi$
    \item $\vec{A} \cup \vec{W}$ is minimal w.r.t. conditions $(1)$ and $(2)$
\end{enumerate}
\textit{AC2} requires that the actual cause \actualcause~should contain only (action) variables for which their underlying conditions (information states) in the counterfactual world $(\mathcal{C}^{\vec{A} \leftarrow \vec{a}', \vec{W} \leftarrow \vec{w}'}, \vec{u})$ are the same as in the actual world $(\mathcal{C}^{}, \vec{u})$. \textit{AC3} says that the contingency $\vec{W} = \vec{w}'$ should contain only variables for which these conditions change. 
Regarding Example \ref{ex.single_bogus}, our definition agrees with the HP definition, because of condition \textit{AC3}. Regarding Example \ref{ex.multi_bogus}, our definition agrees with the BF definition because of condition \textit{AC1}. Furthermore, conditions \textit{AC2} and \textit{AC1} guarantee Properties \ref{prop.ce} and  \ref{prop.min}, respectively.

Definition \ref{def.ours} can also be extended with the notion of normality, by following the same procedure as with the HP definition in \cite{halpern2015graded}.

%% file: 4_responsibility_attribution.tex
\section{Responsibility}\label{sec.resp}

In this section, we study approaches to determining the agents' degree of responsibility relative to a causal setting \causalsetting~and an event of interest $\phi$. We focus on approaches that assign responsibility based on the actual causes of $\phi$ in \causalsetting. More specifically, in Section \ref{sec.chockler-halpern}, we adopt in our setting and analyze a well-known definition of responsibility introduced by Chockler and Halpern \cite{chockler2004responsibility}. In Section \ref{sec.new_def_resp}, we introduce a new family of definitions that extend the Chockler and Halpern definition.
 
\subsection{The Chockler-Halpern Definition}\label{sec.chockler-halpern}

\citet{chockler2004responsibility} show that the original HP definition of causality \cite{halpern2001causes} can be used to assign a degree of responsibility to each primitive event, measuring how pivotal it was for the event of interest. \citet{halpern2016actual} modifies the Chockler-Halpern definition (hereafter CH) to incorporate the other two HP definitions \cite{halpern2005causes, halpern2015modification}.
\citet{alechina2020causality} extend the analysis by appraising the responsibility of each agent.
In our framework and for a definition of causality \acdef, the CH notion of responsibility can be defined as follows.

\begin{definition}\label{def.ch}
(CH)
Consider a causal setting \causalsetting~and an event of interest $\phi$ such that $(\mathcal{C}, \vec{u}) \models \phi$. Agent $i$'s degree of responsibility for $\phi$ in \causalsetting~is $0$ if none of $i$'s actions is a part of an actual cause according to \acdef.
Otherwise it is the maximum value $\frac{m}{k}$ such that if \actualcause~is an actual cause of $\phi$ in \causalsetting~under the contingency $\vec{W} = \vec{w}'$ according to \acdef, then $k = |\vec{A}| + |\vec{W}|$ and $m$ denotes the number of agent $i$'s action variables in $\vec{A}$.
\end{definition}

The CH definition captures the important idea that an agent's degree of responsibility should depend on the size of the actual causes it participates in, the size of their corresponding contingencies, and its degree of participation. However, as mentioned by \citet{baier2021game}, the CH definition does not take into consideration the number of actual causes an agent is involved in, 
which is evidence of that agent's power over the final outcome.
Additionally, the definition also ignores other aspects of actual causality that one might consider important for attributing responsibility, such as the number of different contingencies an actual cause might have.

\subsection{A Family of Methods that Extend CH}\label{sec.new_def_resp}

We consider the CH definition and extend it in a natural way, so that an agent's degree of responsibility is now determined by a wider variety of actual causes, instead of just one. More specifically, our new definition takes into account the whole set of actual cause-witness pairs \acwset~for some definition \acdef~and applies weight vectors over that set. These vectors are non-negative and agent-specific, and they determine by how much an agent's degree of responsibility is affected by each pair in \acwset. Each weight vector $\vec{b}$ has to have at least one strictly positive element.

\begin{definition}\label{def.resp_ours}
(Degree of Responsibility) Consider a causal setting \causalsetting~and an event of interest $\phi$ such that $(\mathcal{C}, \vec{u}) \models \phi$. 
Given a weight vector $\vec{b}$ over the set \acwset, agent $i$'s degree of responsibility for $\phi$ in \causalsetting~is $0$ if none of $i$'s actions is part of an actual cause according to \acdef;
otherwise it is 
\begin{align*}
    \frac{\sum_{c \in \{1, ..., |H_{\mathcal{D}}|\}} b_c \cdot \frac{m_c}{k_c}}{\sum_{c \in \{1, ..., |H_{\mathcal{D}}|\}} b_c}
\end{align*}
such that if \acwpair~is the $c$-th actual cause-witness pair of \acwset, then $k_c = |\vec{A}| + |\vec{W}| - w_c$, and $m_c$ and $w_c$ denote the number of agent $i$'s action variables in $\vec{A}$ and $\vec{W}$, respectively.
\end{definition}

Definition \ref{def.resp_ours} is flexible in the sense that it can generate different responsibility attribution methods by changing the agents' weight vectors.
Compared to Definition \ref{def.ch}, an agent's degree of responsibility does not depend anymore on the number of action variables the agent has in a contingency of an actual cause in which it participates. In simpler words, our definition guarantees that an agent would not be attributed reduced responsibility had it adopted a policy that would make more ``mistakes'' in the counterfactual scenario.\footnote{This guarantee is aligned with the intuition behind the blame attribution property of \textit{Performance Monotonicity}, introduced by \citet{triantafyllou2021blame}.} For instance, in Example \ref{ex.single_bogus}, if $A$'s policy was to \textit{poison} $K$'s water only at time-step $t - 2$ then its degree of responsibility according to CH (and \acdef~being either the HP definition or Definition \ref{def.ours}) would be $1$. However, it would be $1/2$ if its policy was to always \textit{poison} the water. On the contrary, for responsibility attribution methods from Definition \ref{def.resp_ours}, the agent's degree of responsibility is $1$ in both cases.
Note that, similar to CH, an agent's degree of responsibility according to Definition \ref{def.resp_ours} is always between $0$ and $1$.\footnote{The agents' degrees of responsibility do not have to sum up to $1$ \cite{halpern2016actual}.} 
More specifically, if the agent had no impact on the outcome, its degree of responsibility would be $0$, while if it was the only agent with full control over the outcome, its responsibility would be 1.

%% file: 5_experiments.tex
\section{Experiments}\label{sec.experiments}
In this section, we experimentally test the studied definitions of actual causality (Section \ref{sec.ac_definitions}) and responsibility attribution methods (Section \ref{sec.resp}) using an experimental testbed based on the card game Goofspiel. Appendix \ref{sec.add_comp_resuts} contains additional experimental results.

\subsection{Environment and Policies}\label{sec.goofspiel}

\textbf{The game.} \textit{Goofspiel}$(N)$ is a two-person card game where each player's initial hand consists of the cards $\{1, ..., N\}$. There is a face down central pile of cards (also $\{1, ..., N\}$) called the deck, which is shuffled in the beginning of each game. In every round, the top card of the deck is flipped. Then, both players choose a card from their hand and simultaneously reveal it. The player with the higher card wins the round, and in the case of a tie no player wins. If a player manages to win the round, they are awarded a number of points equal to the value of the flipped card, also called the prize, otherwise they are awarded $0$ points. All cards played in that round are then discarded and a new round starts. After $N$ rounds, the player with the most points wins the game. Note that typically $N=13$, making the mathematical analysis of the game quite challenging \cite{ross1971goofspiel, rhoads2012computer}. Moreover, it is worth mentioning that \textit{Goofspiel}$(N)$ is part of a well known framework for RL in games \cite{lanctot2019openspiel, hennes2020neural}.
We introduce a version of this game which has two teams of two players. We call this version \textit{TeamGoofspiel}$(N)$.
The game proceeds as before, with the difference that now the team which cumulatively bids the higher cards in a round is the team that obtains that round's prize. 

\textbf{The players.}
We assume the agency over the members of one of the teams, whose players are referred to as agents and are denoted by \agentzero~and \agentone. We treat the other team as a part of the environment, and we refer to its members as opponents.
All the players are assumed to keep similar information states at each time-step/round. More specifically, the information state of a player at each round consists of: the remaining cards on their hand; the round's prize; partial information about the current score--if their team is winning or not. Notice that for simplicity, in this setting players do not keep track of which cards the other players discarded in previous rounds, i.e., they don't condition their actions on the available moves of other players, nor they try to infer their policies.

\textbf{Policies.} The policy of \agentzero~is to always match the round's prize whenever possible. We differentiate two cases when this action is not feasible, i.e., if the matching card is not on \agentzero's hand. If their team is winning (resp. not winning) they play the card with the highest (resp. lowest) value out of the cards with a value lower (resp. higher) than the prize. In case such a card is not available, they play the card with the lowest (resp. highest) value on their hand.
The policy of \agentone~is to always play the card with the highest value on their hand, if the prize is greater than their hand's average card value minus $X$, otherwise they play the card with the lowest value on their hand.
$X$ is $0$ if their team is winning and $1$ otherwise.

Both opponents follow the same stochastic policy defined as follows. If their team is winning (resp. not winning) they randomly choose a card from their hand with value lower (resp. higher) or equal to the prize, and if they don't have such a card in their hand they randomly choose one of the cards available. 

We choose the players' policies to follow simple rules, and depend on small size information states, so that the generated actual causes are easy to interpret, but not too simple and small, so that they become trivial. Note also that the random and the matching policies are quite standard in \textit{Goofspiel}$(N)$ analysis, and that the latter have been shown to dominate the former \cite{ross1971goofspiel, grimes2013observations}.

\textbf{Actual Causes.} We focus on trajectories, i.e., instances of the game in which the final outcome is either a win for the opponents' team or a draw. For each of these trajectories, our goal is to pinpoint those actions of the agents that caused them to not win the game, and then quantify the agents' influence on that outcome. In particular, we specify the set of all actual cause-witness pairs for each trajectory, and based on this set we compute the agents' degrees of responsibility. Note that in order to generate a trajectory in the \textit{TeamGoofspiel}$(N)$ environment, we first need to sample from the initial state distribution, i.e., shuffle the deck, and then at each time-step sample the opponents' actions based on the distributions defined by their stochastic policies.

\subsection{Demonstration Example}\label{sec.demonstration}

In this section, we focus on a particular trajectory of \textit{TeamGoofspiel}$(N)$ and present the set of actual cause-witness pairs for that trajectory based on Definition \ref{def.ours}. To compute this set, we implement a simple tree search algorithm that iterates over all possible alternative actions of the agents.
In our experiments, we restrict the size of actual cause-witness pairs to $4$, in order to obtain more interpretable actual causes. Namely, large actual causes suggest a counterfactual world that is quite different from what actually happened, and they are also more difficult to comprehend. Thus, smaller actual causes are usually preferred.

For $N=5$, we consider the trajectory where prizes show up in a descending order $(5, 4, 3, 2, 1)$. Both agents' and opponents' actions in this scenario always match the prize, resulting in a $0-0$ draw. Table \ref{tab : ac-w_pairs} shows the set of all actual cause-witness pairs for that trajectory. Interestingly, and despite its simplicity, the considered trajectory admits $19$ different actual cause-witness pairs, each of them representing a set of minimal changes that the agents could have made in order to win the game. More specifically, each row of Table \ref{tab : ac-w_pairs} corresponds to one actual cause-witness pair \acwpair, where: column $1$ includes the actual causes \actualcause; column $2$ includes the counterfactual settings $\vec{a}'$; column $3$ includes the contingencies $\vec{W} = \vec{w}'$; column $4$ includes the improvement in score difference the agents achieve in the corresponding counterfactual worlds.

The corresponding tables for definitions BF and HP can be found in Appendix \ref{sec.demonstration_app}.

\captionsetup{belowskip=0pt}
\begin{table}[t]
\caption{Actual Cause-Witness Pairs Based on Definition \ref{def.ours}}
\label{tab : ac-w_pairs}
\begin{minipage}{\columnwidth}
\begin{center}
\begin{tabular}{|c|c|c|c|}
     \hline
     Actual Cause & CF Setting & Contingency & Improvement \\
     \hline
     $A_{0,1} = 4$, $A_{1,1} = 4$ & $1$, $1$ & - & $2$ \\
     \hline
     \multirow{2}{*}{$A_{0,1} = 4$, $A_{1,3} = 2$} & $1$, $1$ & $A_{0,2} = 2$ & $1$ \\
     \cline{2-4}
     & $2$, $1$ & $A_{0,2} = 1$ & $1$ \\
     \hline
     $A_{0,1} = 4$, $A_{1,2} = 3$ & $1$, $1$ & - & $2$ \\
     \hline
     $A_{1,1} = 4$, $A_{0,3} = 2$ & $1$, $1$ & - & $1$ \\
     \hline
     \multirow{2}{*}{$A_{1,1} = 4$, $A_{0,2} = 3$} & \multirow{2}{*}{$1$, $1$} & $A_{1,2} = 3$ & $2$ \\
     \cline{3-4}
     & & $A_{1,2} = 4$ & $1$ \\
     \hline
     \multirow{2}{*}{$A_{0,0} = 5$} & $1$ & - & $1$ \\
     \cline{2-4}
     & $2$ & - & $1$ \\
     \hline
     \multirow{8}{*}{$A_{1, 0} = 5$} & \multirow{4}{*}{$1$} & $A_{0, 3} = 1$ & $1$\\ 
     \cline{3-4}
     & & $A_{1, 3} = 4$ & $1$ \\
     \cline{3-4}
     & & $A_{1, 1} = 3$ & $3$ \\
     \cline{3-4}
     & & $A_{1, 1} = 4$ & $1$ \\
     \cline{2-4}
     & \multirow{3}{*}{$2$} & $A_{1, 2} = 3$ & $1$ \\
     \cline{3-4}
     & & $A_{1, 1} = 3$ & $3$ \\
     \cline{3-4}
     & & $A_{1, 1} = 4$ & $1$ \\
     \cline{2-4}
     & $3$ & $A_{1, 2} = 2$ & $1$\\
     \hline
     $A_{1, 0} = 5$, $A_{0,2} = 3$ & $1$, $2$ & - & $1$ \\
     \hline
     $A_{1, 0} = 5$, $A_{0,1} = 4$ & $1$, $2$ & - & $1$ \\
     \hline
\end{tabular}
\end{center}
\end{minipage}
\end{table}

\subsection{Instances of Definition \ref{def.resp_ours}}\label{sec.variations}

\begin{figure}[t]
\centering
\includegraphics[width=.7\columnwidth]{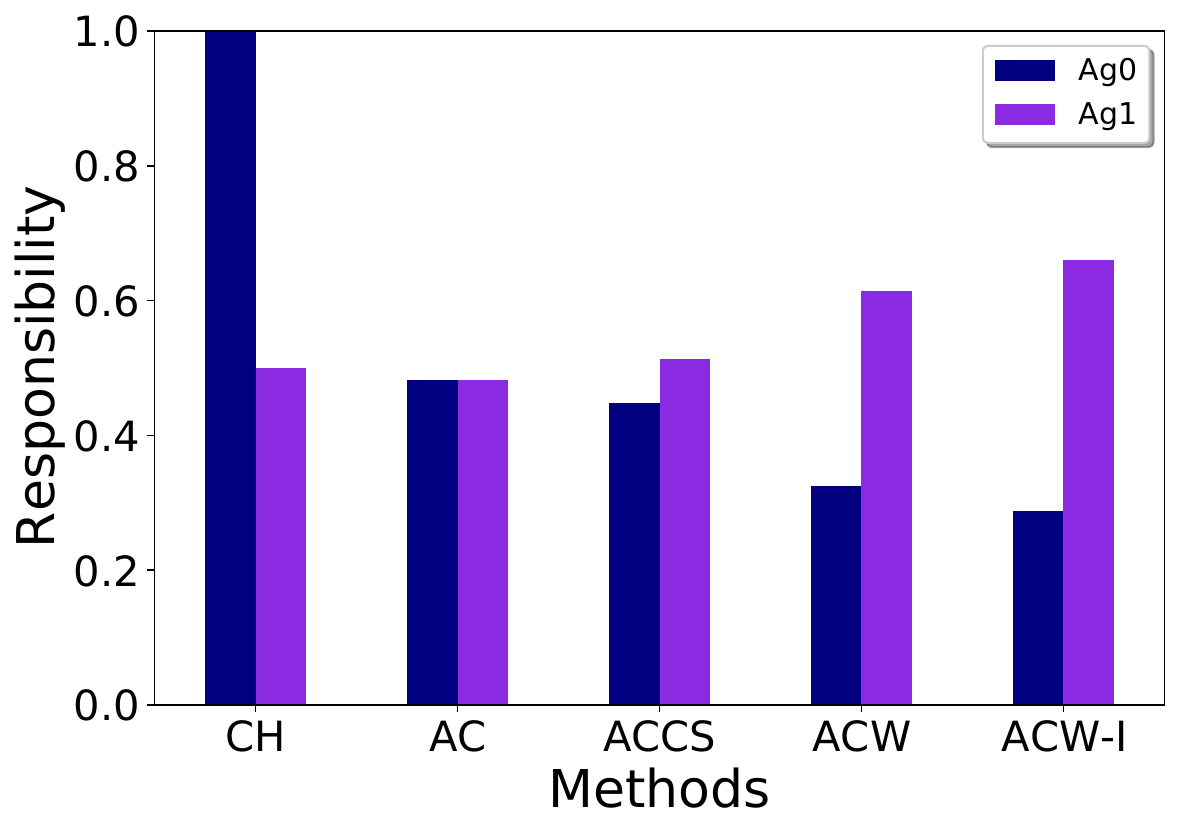}
\captionsetup{type=figure, aboveskip=.5pt, belowskip=-15pt}
\Description{Responsibility Attribution}\
\caption{Responsibility Attribution Based on Table \ref{tab : ac-w_pairs}}
\label{fig: resp_attr}
\end{figure}

As mentioned in Section \ref{sec.new_def_resp}, by changing the values of the agent-specific weight vectors $\vec{b}$ in Definition \ref{def.resp_ours} we can generate multiple different responsibility attribution methods. 
Next, we present the instances of Definition \ref{def.resp_ours} we consider in our experiments. The names of the corresponding attribution methods are derived from the elements of actual causality which they consider distinct.

\textbf{AC:} For every actual cause \actualcause, the weight of exactly one pair \acwpair~is $1$, and all other weights are $0$. For Table \ref{tab : ac-w_pairs}, this means that we keep one row per actual cause, and we discard every other row for that actual cause and the column Improvement. For an agent $i$, the actual cause-witness pair we choose for each actual cause, i.e., the pair whose weight is $1$, is one with the highest value of $\frac{m_c}{k_c}$ for that agent. This method takes into account the number of distinct actual causes an agent participates in.

\textbf{ACCS:} For every actual cause \actualcause~and counterfactual setting $\vec{a}'$, the weight of exactly one pair \acwpair~is $1$, and all other weights are $0$. For Table \ref{tab : ac-w_pairs}, this means that we keep exactly one row per actual cause-counterfactual setting pair, and we discard every other row for that pair and the column Improvement. For an agent $i$, the actual cause-witness pair we choose for each actual cause-counterfactual setting pair, i.e., the pair whose weight is $1$, is one with the highest value of $\frac{m_c}{k_c}$ for that agent. Additional to \textbf{AC}, \textbf{ACCS} takes also into account the number of all the different counterfactual actions that the agents who participate in \actualcause~could have taken in order for the final outcome to improve.

The remaining attribution methods assume the same weight vectors for all agents.

\textbf{ACW:} The weight of each actual cause-witness pair is $1$. For Table \ref{tab : ac-w_pairs}, this means that we take its whole content into account except for the column Improvement. Additional to \textbf{ACCS}, \textbf{ACW} takes into account all the different contingencies under which \actualcause~is an actual cause.

\textbf{ACW-I:} The weight of each actual cause-witness pair is equal to the value of the counterfactual improvement it admits. This means that we use the full information given in this table.\footnote{This approach to responsibility is aligned with the notion of graded causality \cite{halpern2015graded}. However, here instead of using a normality ranking over the actual cause-witness pairs, we evaluate them w.r.t. the counterfactual improvement they admit.}

Apart from \textbf{AC}, \textbf{ACCS}, \textbf{ACW} and \textbf{ACW-I}, we also consider in our experiments the CH definition. Plot \ref{fig: resp_attr} shows the agents' degrees of responsibility for the trajectory from Section \ref{sec.demonstration} and for the various responsibility attribution methods. For this plot, the input of all the methods is Table~\ref{tab : ac-w_pairs}. Observe how in this example the lion's share of responsibility shifts gradually from \agentzero~to \agentone, as we include more information from Table \ref{tab : ac-w_pairs} to our responsibility assignment process. For instance, \agentone~could improve the outcome on their own by playing one of $3$ alternative actions at the first time-step (rows $10$-$16$), while \agentzero~had only $2$ (rows $8$, $9$). Because of that, \agentone's responsibility increases relative to \agentzero's when we transition from \textbf{AC} to \textbf{ACCS}. 
Appendix \ref{sec.variations_app} shows the attributed responsibilities, when BF and HP are considered instead of Definition \ref{def.ours}.


\subsection{Violations of Properties \ref{prop.ce} and  \ref{prop.min}}\label{sec.violations}

In this section, we compute the frequency of Property \ref{prop.ce} and Property \ref{prop.min} violations by the BF and HP definitions from Section \ref{sec.ac_definitions}. Furthermore, we examine by how much these property violations might affect the agents' degrees of responsibility. We measure both quantities for $N \in \{4, 5, 6, 7, 8\}$, and $50$ trajectories per value of $N$.

\captionsetup[figure]{belowskip=0pt}
\begin{figure*}
\captionsetup[subfigure]{aboveskip=0pt,belowskip=1pt}
\centering
    \begin{subfigure}[c]{0.32\textwidth}
        \includegraphics[width=\textwidth, height=0.195\textheight]{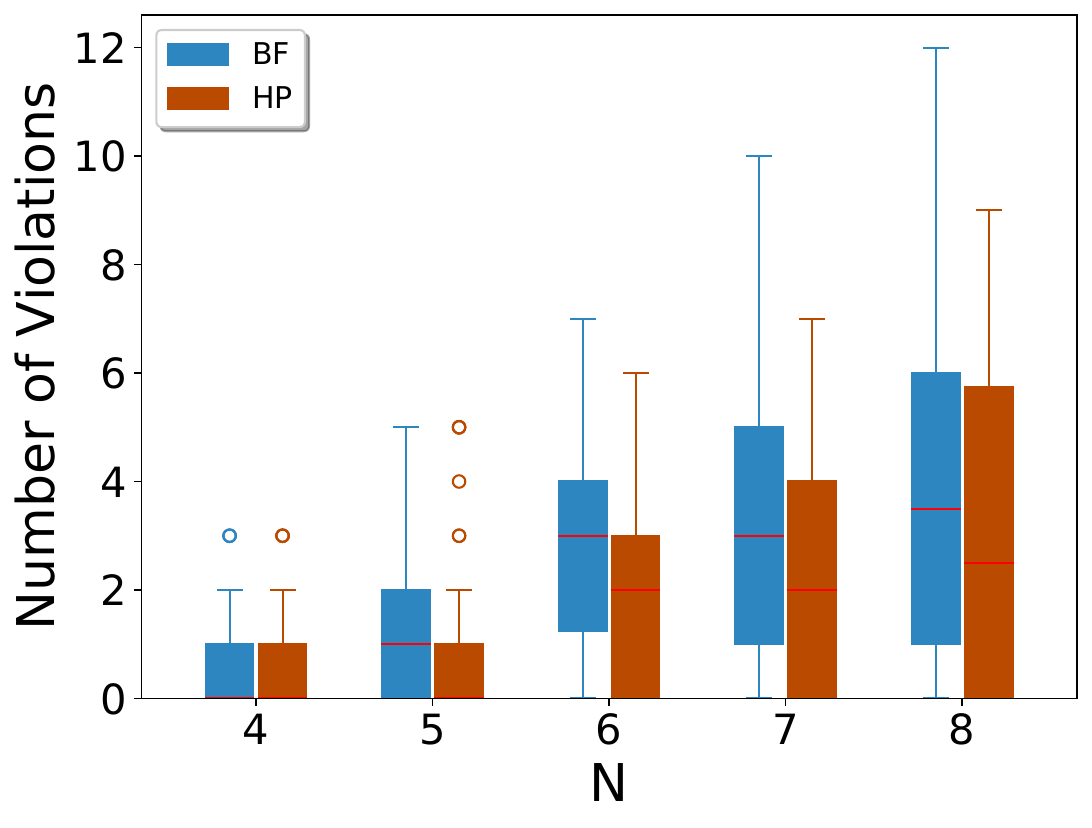}
        \captionsetup{type=figure}
        \caption{Property \ref{prop.ce}: Number of Violations}
        \label{fig : prop1_violations}
    \end{subfigure}\hfill\hfill%
    \begin{subfigure}[c]{0.64\textwidth}
        \includegraphics[width=\textwidth, height=0.195\textheight]{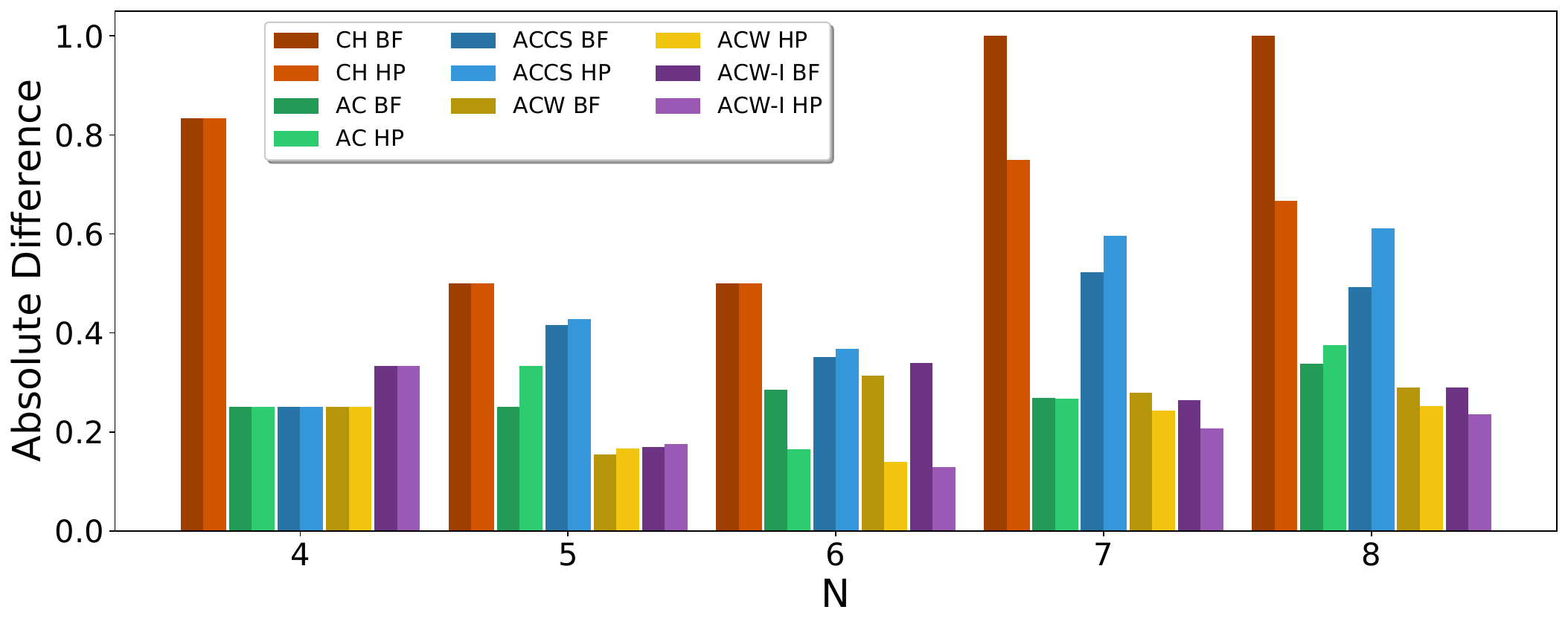}
        \captionsetup{type=figure}
        \caption{Property \ref{prop.ce}: Impact on Responsibility}
        \label{fig : prop1_resp}
    \end{subfigure}\\
    \begin{subfigure}[c]{0.32\textwidth}
        \includegraphics[width=\textwidth, height=0.195\textheight]{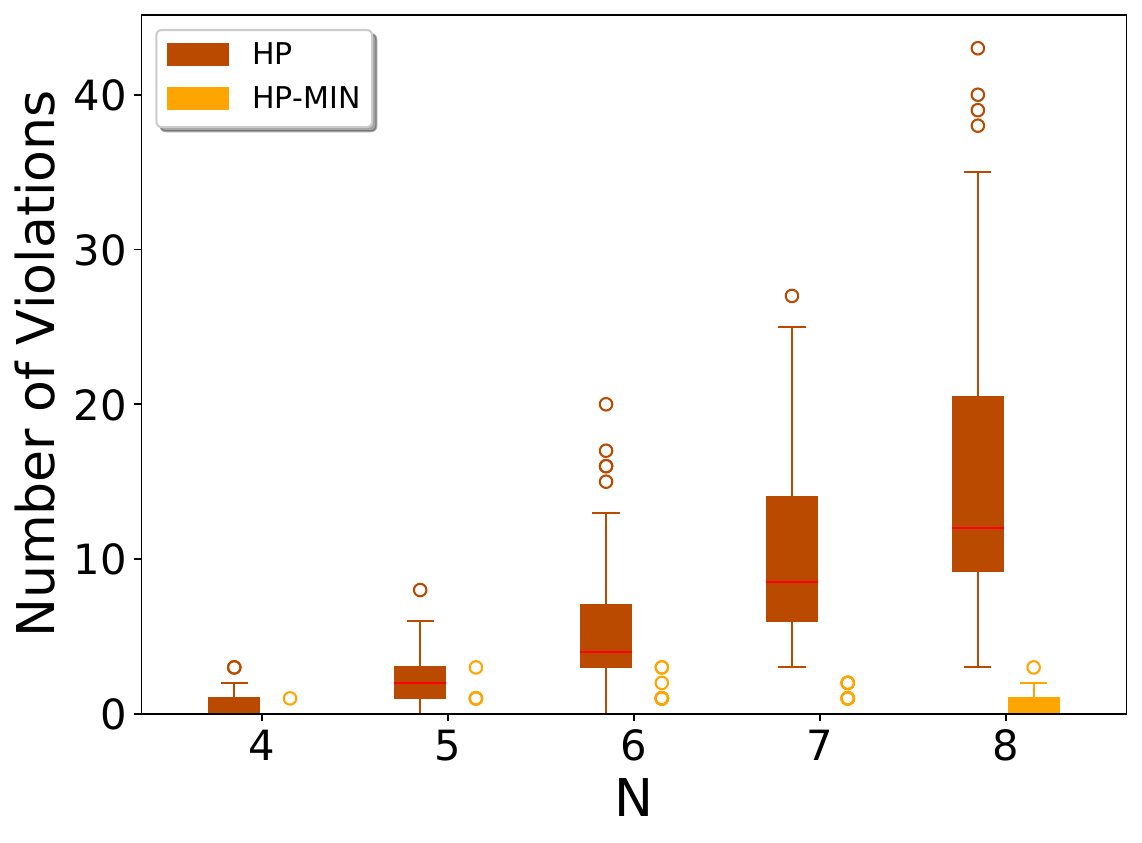}
        \captionsetup{type=figure}
        \caption{Property \ref{prop.min}: Number of Violations}
        \label{fig : prop3_violations}
    \end{subfigure}\hfill%
    \begin{subfigure}[c]{0.32\textwidth}
        \includegraphics[width=\textwidth, height=0.195\textheight]{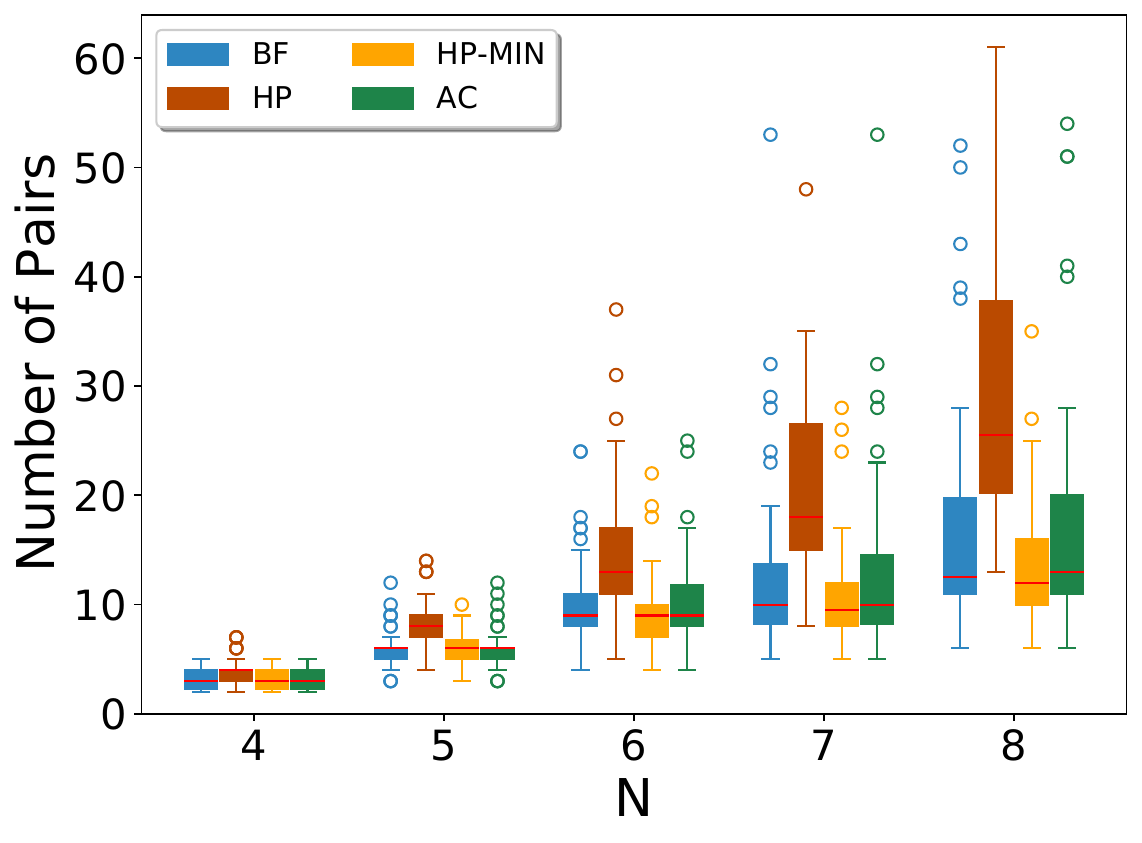}
        \captionsetup{type=figure}
        \subcaption{Actual Cause-Contingency Pairs}
        \label{fig : num_ac_cont}
    \end{subfigure}\hfill%
    \begin{subfigure}[c]{0.32\textwidth}
        \includegraphics[width=\textwidth, height=0.195\textheight]{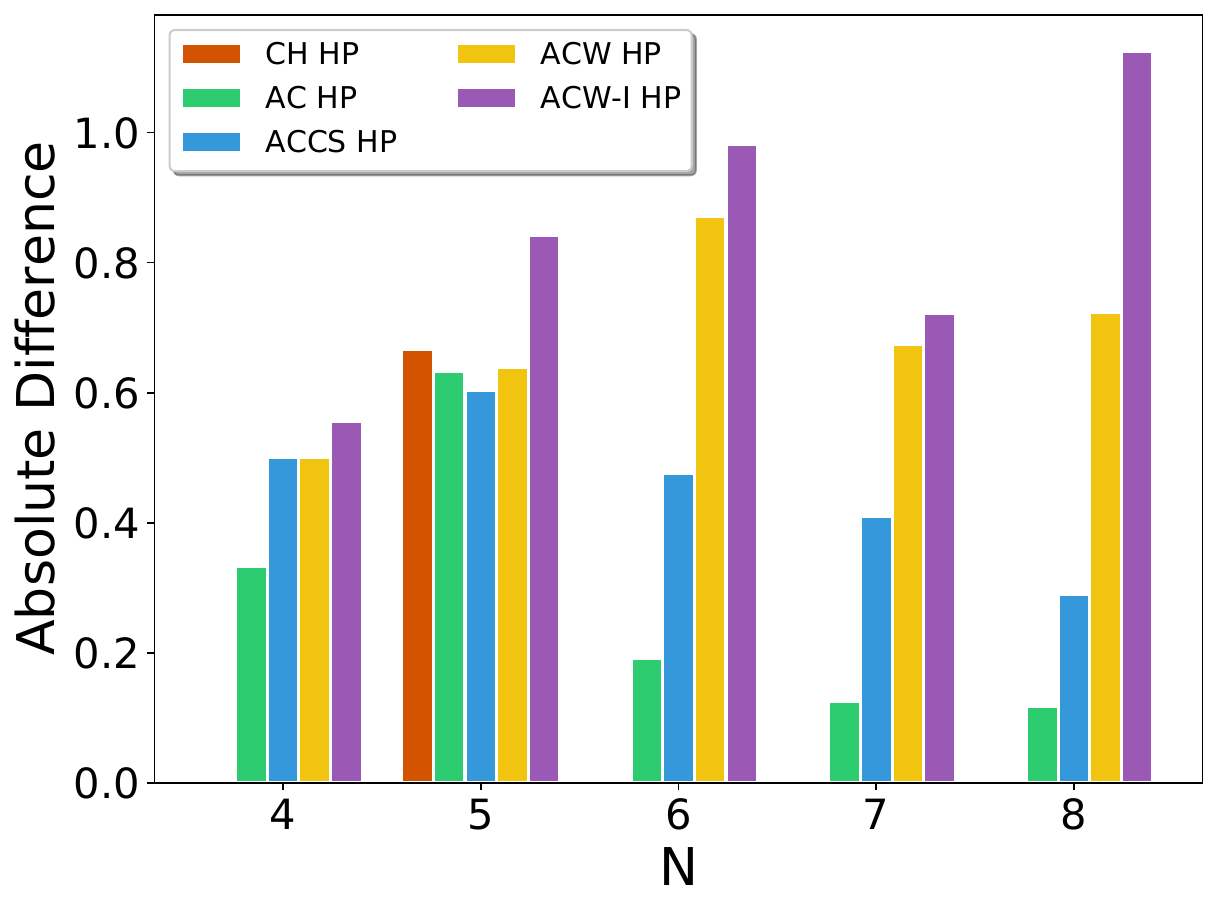}
        \captionsetup{type=figure}
        \caption{Property \ref{prop.min}: Impact on Responsibility}
        \label{fig : prop3_resp}
    \end{subfigure}
\Description{Plots \ref{fig : prop1_violations} and \ref{fig : prop3_violations} show the number of violations of Properties \ref{prop.ce} and \ref{prop.min}. Plots \ref{fig : prop1_resp} and \ref{fig : prop3_resp} show the impact of these violations on the agents' degrees of responsibility. Plot \ref{fig : num_ac_cont} shows the number of distinct actual cause-contingency pairs. All plots are shown over $50$ trajectories per value of $N$.}
\caption{Plots \ref{fig : prop1_violations} and \ref{fig : prop3_violations} show the number of violations of Properties \ref{prop.ce} and \ref{prop.min}. Plots \ref{fig : prop1_resp} and \ref{fig : prop3_resp} show the impact of these violations on the agents' degrees of responsibility. Plot \ref{fig : num_ac_cont} shows the number of distinct actual cause-contingency pairs.} 
\label{fig: main_plots}
\end{figure*}

\subsubsection{Property \ref{prop.ce} Violations}\label{sec.prop1_violations}
Plot \ref{fig : prop1_violations} summarizes the frequency results for Property \ref{prop.ce}. More specifically, for each trajectory we compute the number of actions that are, according to Property \ref{prop.ce}, wrongfully characterized as part of one or more actual causes by the BF and HP definitions. For instance, consider the boxplot which corresponds to $N = 6$ and HP. For half of the trajectories, the HP definition considers at least $2$ out of the total $12$ actions as part of one or more actual causes, when it should had instead considered them as part of their contingencies.

Next, we want to measure by how much this mislabeling of actions, i.e., Property \ref{prop.ce} violations, can affect the process of responsibility attribution. In order to quantify this measure, we execute the following procedure. For each trajectory, we first compute the set of actual cause-witness pairs \acwset~based on definition \acdef, where \acdef~can be either BF or HP. Then, for every approach from Section \ref{sec.variations} we compute the agents' degrees of responsibility utilizing the set \acwset. Next, we correct \acwset~for Property \ref{prop.ce} violations, i.e., for every actual cause-witness pair \acwpair~in \acwset~we remove from \actualcause~and $\vec{a}'$ all actions that violate Property \ref{prop.ce}, and we add them to the contingency set $\vec{W} = \vec{w}'$. We then take the newly defined set of actual cause-witness pairs $H'_{\mathcal{D}}$, and recompute the agents' degrees of responsibility. 
Plot \ref{fig : prop1_resp} shows the maximum value of the total absolute difference between the two computed degrees of responsibility, for every value of $N$ and responsibility method. The maximum is taken over all trajectories.
We choose to plot the maximum differences to showcase the potential magnitude of unfairness that violating Property \ref{prop.ce} might cause to the responsibility assignment. The results demonstrate that correcting BF and HP for these violations can have a significant impact on the agents' degrees of responsibility. It is also worth noting that the CH definition seems to be the least resilient to this type of violations among the definitions we consider here.

\subsubsection{Property \ref{prop.min} Violations}\label{sec.prop3_violations}
Apart from Property \ref{prop.ce}, the HP definition also violates Property \ref{prop.min} (Section \ref{sec.hp_def}). Plot \ref{fig : prop3_violations} displays the frequency of these violations (brown boxplots). More specifically, the plot shows for all trajectories the number of distinct actual cause-contingency pairs which are non-minimal, according to Property \ref{prop.min}. While comparing this number to the total number of these pairs which is shown in Plot \ref{fig : num_ac_cont}, we conclude that the HP definition systematically violates Property \ref{prop.min}.

To measure the impact of Property \ref{prop.min} violations on responsibility attribution, we follow a procedure similar to the one for Property \ref{prop.ce} in Section \ref{sec.prop1_violations}. More specifically, we first compute the agents' degrees of responsibility based on $H_\mathcal{D}$, where \acdef~is the HP definition. Next, we correct $H_\mathcal{D}$ for Property \ref{prop.min} violations, i.e., we remove all actual cause-witness pairs that violate Property \ref{prop.min}, and we recompute the degrees of responsibility. 
Similar to Plot \ref{fig : prop1_resp}, Plot \ref{fig : prop3_resp} shows the maximum value of the total absolute difference between the two computed degrees of responsibility, for every value of $N$ and responsibility method. The maximum is taken over all trajectories.
The results indicate that Property \ref{prop.min} violations in the HP definition can greatly affect the downstream task of responsibility attribution.
However, it is worth mentioning that for CH, the agents' degrees of responsibility changed only for $1$ trajectory out of the $250$ we sampled in our experiments, indicating that it is the most resilient to this type of property violations.

Note that the HP definition allows for non-minimal contingencies, that is \acwpair~may be considered as a valid actual cause-witness pair by HP even if there is a subset $\vec{W}'$ of $\vec{W}$ and a setting $\vec{w}''$ such that $(\vec{A} = \vec{a}, (\vec{W}', \vec{w}'', \vec{a}'))$ is also an actual cause-witness pair according to HP. As mentioned by \citet{ibrahim2021actual}, when attributing responsibility based on the HP definition it would make sense to impose the witness minimality condition in addition to \textit{HPC3}. We denote this enhanced version of the HP definition by HP-MIN. 
Note that violations of the contingency minimality condition fall under Property \ref{prop.min} violations. Therefore, we expect that HP-MIN will do better than HP w.r.t Property \ref{prop.min} violations, and hence have a lower impact on responsibility.\footnote{In particular, only \textbf{ACCS}, \textbf{ACW} and \textbf{ACW-I} are affected.} Plots \ref{fig : prop3_violations} and \ref{fig : num_ac_cont} verify this intuition. They show that the number of violations is significantly smaller for HP-MIN. However, these violations are not completely vanished, meaning that there are still cases where they can have a large impact on responsibility attribution. In Appendix \ref{sec.hp-min}, we plot again \ref{fig : prop1_violations}, \ref{fig : prop1_resp} and \ref{fig : prop3_resp}, after replacing HP with HP-MIN.

%% file: 6_conclusion.tex
\section{Conclusion}

To summarize, in this paper we studied actual causality and responsibility attribution through the lens of sequential decision making in Dec-POMDs. We identified some of the shortcomings of existing definitions of actual causality and introduced a new definition to address them. Furthermore, we extended one of the well known causality-based approaches to responsibility attribution 
in order to account for an agent's power over the final outcome and its ability to manipulate its own degree of responsibility.
While this work focuses on particular challenges in defining actual causality and attributing responsibility, we view it as an important step toward establishing a formal framework that supports accountable multi-agent sequential decision making.

Some of the most interesting future research directions are related to practical considerations. Given that our primary goal is to formalize the notions of actual causality and responsibility attribution, we made simplifying assumptions that allowed more efficient computation of experimental results. For example, even though the Dec-POMDP framework adopted in this work does model uncertainty, we assumed the full knowledge of random variables that define contexts of the corresponding SCM. We also assumed that the agents' policies are given. Lifting these assumptions is critical for making this work more applicable in practice.
Furthermore, the algorithmic solution for inferring actual causes and assign responsibility in the experiments is based on exhaustive search. Therefore, deriving more scalable algorithmic solutions is needed for applying this work in challenging domains. 
Finally, we deem further analysis of actual causality properties a meaningful extension of our work.

%% file: 8_acknowledgements.tex
\section*{Acknowledgements}
This research was, in part, funded by the Deutsche Forschungsgemeinschaft (DFG, German Research Foundation) – project number $467367360$. We thank the anonymous reviewers for their valuable comments and suggestions. 

%% file: 7.1_causal_graph.tex

\section{Causal Graph of Dec-POMDP SCM}\label{sec.dag}

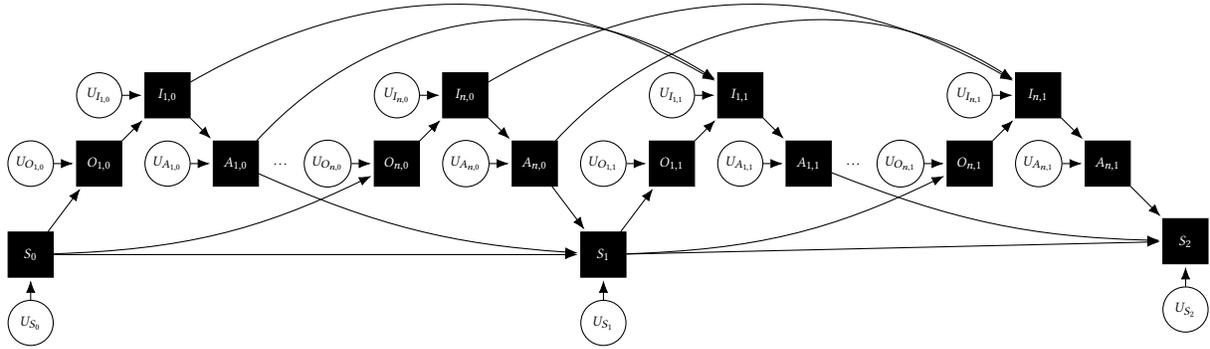
\begin{figure*}
\centering
\begin{tikzpicture}[>=stealth', shorten >=1pt, auto,
    node distance=0.5cm, scale=.6, 
    transform shape, align=center, 
    state/.style={circle, draw, minimum size=1cm}]


\node[state] (UO-10) at (0,0)  {$U_{O_{1,0}}$};
\node[state, rectangle, fill=black, text=white] (O-10) [right=of UO-10] {$O_{1,0}$};
\path (UO-10) edge (O-10);

\node[state] (UI-10) [above=of O-10] {$U_{I_{1,0}}$};
\node[state, rectangle, fill=black, text=white] (I-10) [right=of UI-10] {$I_{1,0}$};
\path (UI-10) edge (I-10);
\path (O-10) edge (I-10);

\node[state] (UA-10) [right=of O-10] {$U_{A_{1,0}}$};
\node[state, rectangle, fill=black, text=white] (A-10) [right=of UA-10] {$A_{1,0}$};
\path (UA-10) edge (A-10);
\path (I-10) edge (A-10);

\node[state] (UO-n0) [right=1cm of A-10] {$U_{O_{n,0}}$};
\node[state, rectangle, fill=black, text=white] (O-n0) [right=of UO-n0] {$O_{n,0}$};
\path (UO-n0) edge (O-n0);

\path (A-10) -- node[auto=false]{\ldots} (UO-n0);

\node[state] (UI-n0) [above=of O-n0] {$U_{I_{n,0}}$};
\node[state, rectangle, fill=black, text=white] (I-n0) [right=of UI-n0] {$I_{n,0}$};
\path (UI-n0) edge (I-n0);
\path (O-n0) edge (I-n0);

\node[state] (UA-n0) [right=of O-n0] {$U_{A_{n,0}}$};
\node[state, rectangle, fill=black, text=white] (A-n0) [right=of UA-n0] {$A_{n,0}$};
\path (UA-n0) edge (A-n0);
\path (I-n0) edge (A-n0);

\node[state, rectangle, fill=black, text=white] (S-0) [below=1cm of UO-10] {$S_0$};
\node[state] (US-0) [below=of S-0] {$U_{S_0}$};
\path (US-0) edge (S-0);
\path (S-0) edge (O-10);
\path (S-0) edge[bend right=12]  (O-n0);


\node[state] (UO-11) [right=of A-n0]  {$U_{O_{1,1}}$};
\node[state, rectangle, fill=black, text=white] (O-11) [right=of UO-11] {$O_{1,1}$};
\path (UO-11) edge (O-11);

\node[state] (UI-11) [above=of O-11] {$U_{I_{1,1}}$};
\node[state, rectangle, fill=black, text=white] (I-11) [right=of UI-11] {$I_{1,1}$};
\path (UI-11) edge (I-11);
\path (O-11) edge (I-11);
\path (I-10)[bend left=30] edge (I-11);
\path (A-10)[bend left=40] edge (I-11);

\node[state] (UA-11) [right=of O-11] {$U_{A_{1,1}}$};
\node[state, rectangle, fill=black, text=white] (A-11) [right=of UA-11] {$A_{1,1}$};
\path (UA-11) edge (A-11);
\path (I-11) edge (A-11);

\node[state] (UO-n1) [right=1cm of A-11] {$U_{O_{n,1}}$};
\node[state, rectangle, fill=black, text=white] (O-n1) [right=of UO-n1] {$O_{n,1}$};
\path (UO-n1) edge (O-n1);

\path (A-11) -- node[auto=false]{\ldots} (UO-n1);

\node[state] (UI-n1) [above=of O-n1] {$U_{I_{n,1}}$};
\node[state, rectangle, fill=black, text=white] (I-n1) [right=of UI-n1] {$I_{n,1}$};
\path (UI-n1) edge (I-n1);
\path (O-n1) edge (I-n1);
\path (I-n0)[bend left=30] edge (I-n1);
\path (A-n0)[bend left=40] edge (I-n1);

\node[state] (UA-n1) [right=of O-n1] {$U_{A_{n,1}}$};
\node[state, rectangle, fill=black, text=white] (A-n1) [right=of UA-n1] {$A_{n,1}$};
\path (UA-n1) edge (A-n1);
\path (I-n1) edge (A-n1);

\node[state, rectangle, fill=black, text=white] (S-1) [below=1cm of UO-11] {$S_1$};
\node[state] (US-1) [below=of S-1] {$U_{S_1}$};
\path (US-1) edge (S-1);
\path (S-0) edge (S-1);
\path (S-1) edge (O-11);
\path (S-1) edge[bend right=12]  (O-n1);
\path (A-10) edge[bend right=10] (S-1);
\path (A-n0) edge (S-1);

\node[state, rectangle, fill=black, text=white] (S-2) [below right=1cm of A-n1] {$S_2$};
\node[state] (US-2) [below=of S-2] {$U_{S_2}$};
\path (US-2) edge (S-2);
\path (S-1) edge (S-2);
\path (A-11) edge[bend right=10] (S-2);
\path (A-n1) edge (S-2);

\end{tikzpicture}

\captionsetup{type=figure}
\Description{Dec-POMDP in the view of SCM}
\caption{Causal Graph of Dec-POMDP SCM with Structural Equations \eqref{eq.struct_eq}}
\label{fig: dec_scm}
\end{figure*}

In this section, we provide the causal graph of the Dec-POMDP SCM from Section \ref{sec.decpomdp_scm}. The graph is shown in Figure \ref{fig: dec_scm}.

%% file: 7.2_gumbel.tex

\section{Gumbel-Max SCMs and Counterfactual Stability}\label{sec.gumbel}

In this section, we show how Gumbel-Max SCMs can be implemented in our framework, and also provide the main intuition behind the counterfactual stability property. For more details on Gumbel-Max SCMs and the counterfactual stability property we refer the interested reader to \cite{oberst2019counterfactual}.

Under the Gumbel-Max model, the structural equations of Eq. \eqref{eq.struct_eq} become:
\begin{align*}\label{eq.gum_struct_eq}
    & S_t = \argmax_{s \in \mathcal{S}}\{\log P(S_{t-1}, A_{t-1}, S_t = s) + U_{S_t}\}\\
    & O_t = \argmax_{o \in \mathcal{O}}\{\log \Omega(S_t, O_t = o) + U_{O_t}\}\\
    & I_{i,t} = \argmax_{\imath_i \in \mathcal{I}_i}\{\log Z_i(I_{i,t-1}, A_{i,t-1}, O_t, I_{i,t} = \imath_i) + U_{I_{i,t}}\}\\
    & A_{i,t} = \argmax_{a_i \in \mathcal{A}_i}\{\log \pi_i(A_{i, t} = a_i|I_{i,t}) + U_{A_{i,t}}\}
\end{align*}
where  $U_{S_t}$, $U_{O_t}$, $U_{I_{i,t}}$ and $U_{A_{i,t}} \sim$ Gumbel$(0, 1)$. 

The class of Gumbel-Max SCMs has been shown to satisfy the desirable property of counterfactual stability, which excludes a specific type of non-intuitive counterfactual outcomes. We provide the main intuition behind this property, with the help of an example. Consider the observed trajectory $\tau = \{(s_t, a_t)\}_{t=0}^{T-1}$, and the counterfactual scenario in which agents $\{1, ..., n\}$ take the joint action $a'$ at time-step $t$, instead of $a_t$. The counterfactual stability property ensures that under this counterfactual scenario, it is impossible that at time-step $t+1$ the process would transition to a state $s'$ different than the observed state, i.e., $s_{t+1}$ if
\begin{align*}
    \frac{P(S_t = s, A_t = a', S_{t+1} = s_{t+1})}{P(S_t = s, A_t = a_t, S_{t+1} = s_{t+1})} \ge 
    \frac{P(S_t = s, A_t = a', S_{t+1} = s')}{P(S_t = s, A_t = a_t, S_{t+1} = s')}.
\end{align*}
In other words, in order for the state at time-step $t+1$ to change under a counterfactual scenario, the relative likelihood of an alternative state $s'$ must have increased compared to that of $s_{t+1}$.

%% file: 7.3_demo.tex

\section{Results of Sections \ref{sec.demonstration} and \ref{sec.variations} for the BF and HP definitions}

\subsection{Actual Cause-Witness Pairs}\label{sec.demonstration_app}

In this section, we provide the sets of actual cause-witness pairs for the trajectory considered in Section \ref{sec.demonstration}, and definitions BF (Table \ref{tab : ac-w_pairs_bf}) and HP (Table \ref{tab : ac-w_pairs_hp}).

We also describe a scenario where the HP definition violates Property \ref{prop.ce}. The actual cause-witness pair of Table \ref{tab : ac-w_pairs_hp} denoted by red color fails to meet the conditions stated by Property \ref{prop.ce}. More specifically, \agentzero's information state at time-step $2$ is different between the actual world and the witness world, i.e., the agent's hand in the actual situation at time $2$ is $(1,2,3)$ and in the counterfactual scenario is $(2,3,4)$. Despite that, action $A_{0,2} = 4$ is still considered as a part of the actual cause by the HP definition. Therefore, this scenario shows that the HP definition does not satisfy Property \ref{prop.ce}.


\captionsetup{belowskip=0pt}
\begin{table*}
\caption{Actual Cause-Witness Pairs Based on the BF Definition}
\label{tab : ac-w_pairs_bf}
\begin{minipage}{\textwidth}
\begin{center}
\begin{tabular}{|c|c|c|c|}
     \hline
     Actual Cause & CF Setting & Contingency & Improvement \\
     \hline
     $A_{0,1} = 4$, $A_{1,1} = 4$ & $1$, $1$ & - & $2$ \\%
     \hline
     \multirow{2}{*}{$A_{0,1} = 4$, $A_{0,2} = 3$, $A_{1,3} = 2$} & $1$, $2$, $1$ & - & $1$ \\ %
     \cline{2-4}
     & $2$, $1$,$1$ & - & $1$ \\%
     \hline
     $A_{0,1} = 4$, $A_{1,2} = 3$ & $1$, $1$ & - & $2$ \\ %
     \hline
     $A_{1,1} = 4$, $A_{0,3} = 2$ & $1$, $1$ & - & $1$ \\ %
     \hline
     \multirow{2}{*}{$A_{1,1} = 4$, $A_{0,2} = 3$, $A_{1,2} = 3$} & $1$, $1$, $3$ & - & $2$ \\ %
     \cline{2-4}
     & $1$, $1$, $4$ & - & $1$ \\ %
     \hline
     \multirow{2}{*}{$A_{0,0} = 5$} & $1$ & - & $1$ \\ %
     \cline{2-4}
     & $2$ & - & $1$ \\ %
     \hline
     $A_{1, 0} = 5$, $A_{0, 3} = 2$ & $1$, $1$ & - & $1$\\ %
     \hline
     $A_{1, 0} = 5$, $A_{1, 3} = 2$ & $1$, $4$ & - & $1$ \\ %
     \hline
     \multirow{4}{*}{$A_{1, 0} = 5$, $A_{1, 1} = 4$} & $1$, $3$ & - & $3$ \\ %
     \cline{2-4}
     & $1$, $4$ & - & $1$ \\ %
     \cline{2-4}
     & $2$, $3$ & - & $3$ \\ %
     \cline{2-4}
     & $2$, $4$ & - & $1$ \\ %
     \hline
     \multirow{2}{*}{$A_{1, 0} = 5$, $A_{1, 2} = 3$} & $2$, $3$ & - & $1$ \\ %
     \cline{2-4}
     & $3$, $2$ & - & $1$ \\ %
     \hline
     $A_{1, 0} = 5$, $A_{0,2} = 3$ & $1$, $2$ & - & $1$ \\ %
     \hline
     $A_{1, 0} = 5$, $A_{0,1} = 4$ & $1$, $2$ & - & $1$ \\ %
     \hline
\end{tabular}
\end{center}
\end{minipage}
\end{table*}
\captionsetup{belowskip=0pt}
\begin{table*}
\caption{Actual Cause-Witness Pairs Based on the HP Definition}
\label{tab : ac-w_pairs_hp}
\begin{minipage}{\textwidth}
\begin{center}
\begin{tabular}{|c|c|c|c|}
     \hline
     Actual Cause & CF Setting & Contingency & Improvement \\
     \hline
     $A_{0,1} = 4$, $A_{1,1} = 4$ & $1$, $1$ & - & $2$ \\%
     \hline
     \multirow{2}{*}{\textcolor{red}{$A_{0,1} = 4$, $A_{0,2} = 3$, $A_{1,3} = 2$}} & \textcolor{red}{$1$, $2$, $1$} & \textcolor{red}{-} & \textcolor{red}{$1$} \\%
     \cline{2-4}
     & $2$, $1$, $1$ & - & $1$ \\%
     \hline
     $A_{0,1} = 4$, $A_{1,2} = 3$ & $1$, $1$ & - & $2$ \\%
     \hline
     $A_{1,1} = 4$, $A_{0,3} = 2$ & $1$, $1$ & - & $1$ \\%
     \hline
     $A_{1,1} = 4$, $A_{0,2} = 3$ & $1$, $1$ & $A_{1,2} = 3$ & $2$ \\%
     \hline
     \multirow{2}{*}{$A_{0,0} = 5$} & $1$ & - & $1$ \\%
     \cline{2-4}
     & $2$ & - & $1$ \\%
     \hline
     \multirow{5}{*}{$A_{1, 0} = 5$} & \multirow{2}{*}{$1$} & $A_{1, 1} = 4$ & $1$ \\%
     \cline{3-4}
     & & $A_{1, 1} = 4$, $A_{1, 2} = 3$ & $1$ \\%
     \cline{2-4}
     & \multirow{3}{*}{$2$} & $A_{1, 2} = 3$ & $1$ \\%
     \cline{3-4}
     & & $A_{1, 1} = 4$ & $1$ \\%
     \cline{3-4}
     & & $A_{1, 1} = 4$, $A_{1, 2} = 3$ & $1$ \\%
     \hline
\end{tabular}
\end{center}
\end{minipage}
\end{table*}

\subsection{Responsibility}\label{sec.variations_app}

\begin{figure*}
\centering
\begin{minipage}{0.48\textwidth}
    \centering
    \includegraphics[width=.7\columnwidth]{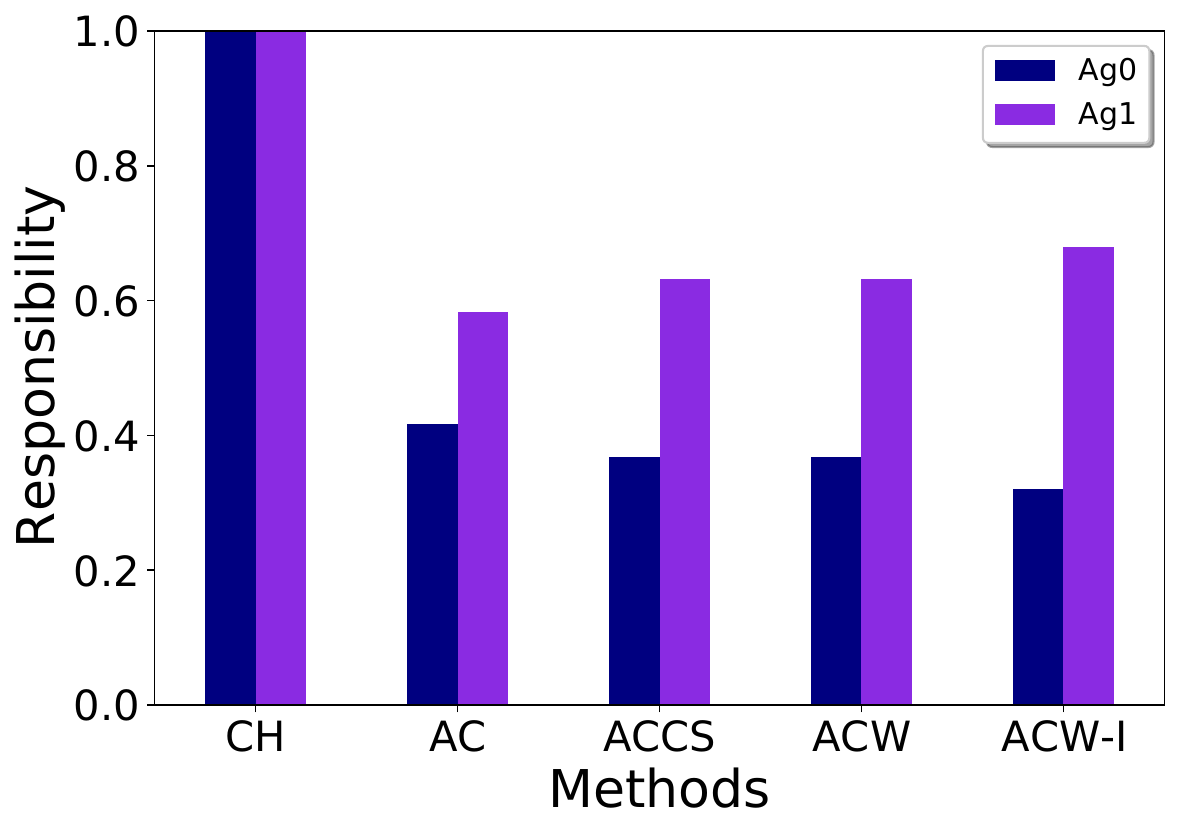}
    \captionsetup{type=figure}
    \caption{Responsibility Attribution Based on Table \ref{tab : ac-w_pairs_bf}}
    \label{fig: resp_attr_bf}
\end{minipage}
\begin{minipage}{0.48\textwidth}
    \centering
    \includegraphics[width=.7\columnwidth]{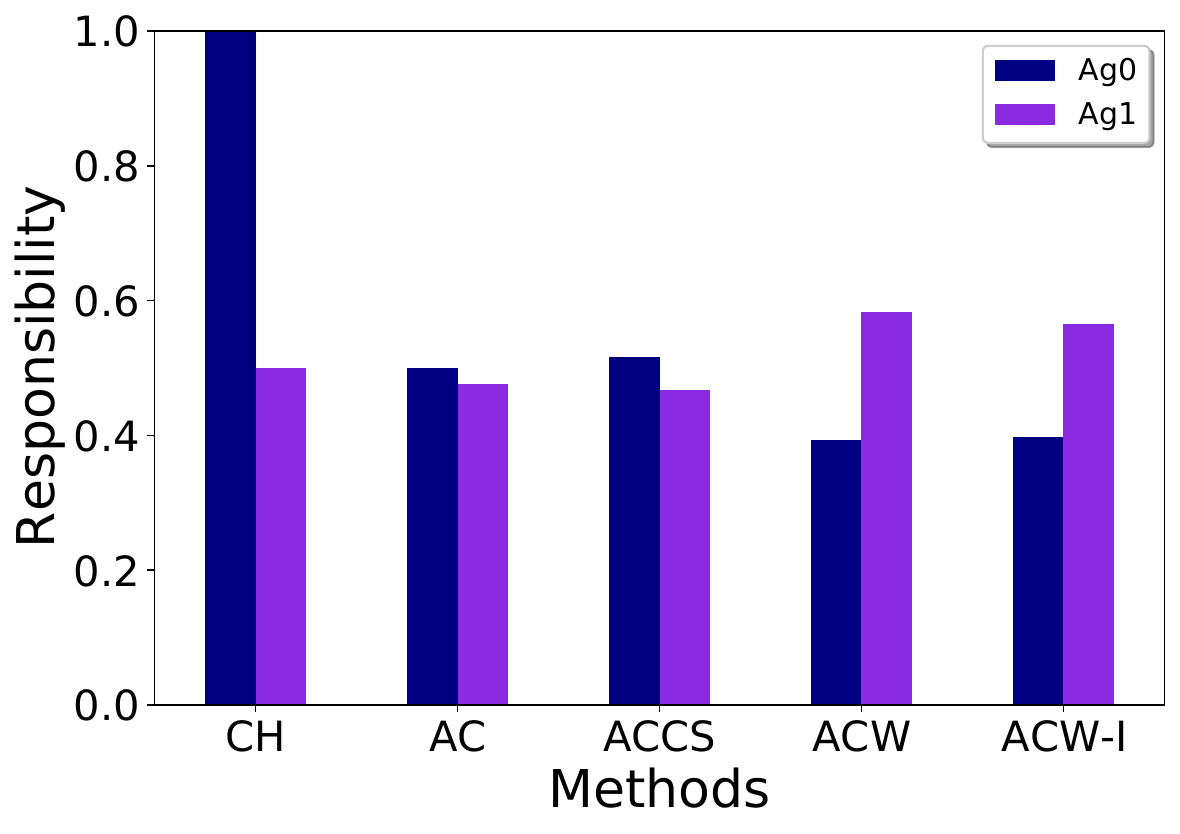}
    \captionsetup{type=figure}
    \caption{Responsibility Attribution Based on Table \ref{tab : ac-w_pairs_hp}}
    \label{fig: resp_attr_hp} 
\end{minipage}
\end{figure*}

In this section, we provide the agents' degrees of responsibility for the trajectory considered in Section \ref{sec.demonstration}, and definitions BF (Plot \ref{fig: resp_attr_bf}) and HP (Plot \ref{fig: resp_attr_hp}). Compared to Plot \ref{fig: resp_attr}, Plots \ref{fig: resp_attr_bf} and \ref{fig: resp_attr_hp} show a similar albeit less prominent tendency, regarding the shift of responsibility from \agentzero~to \agentone, throughout the several approaches to responsibility attribution considered in this paper.


%% file: 7.4_hp-min.tex

\section{Experimental Results for HP-MIN}\label{sec.hp-min}

\captionsetup[figure]{belowskip=0pt}
\begin{figure*}
\captionsetup[subfigure]{aboveskip=0pt,belowskip=1pt}
    \begin{subfigure}[c]{0.32\textwidth}
        \includegraphics[width=\textwidth, height=0.195\textheight]{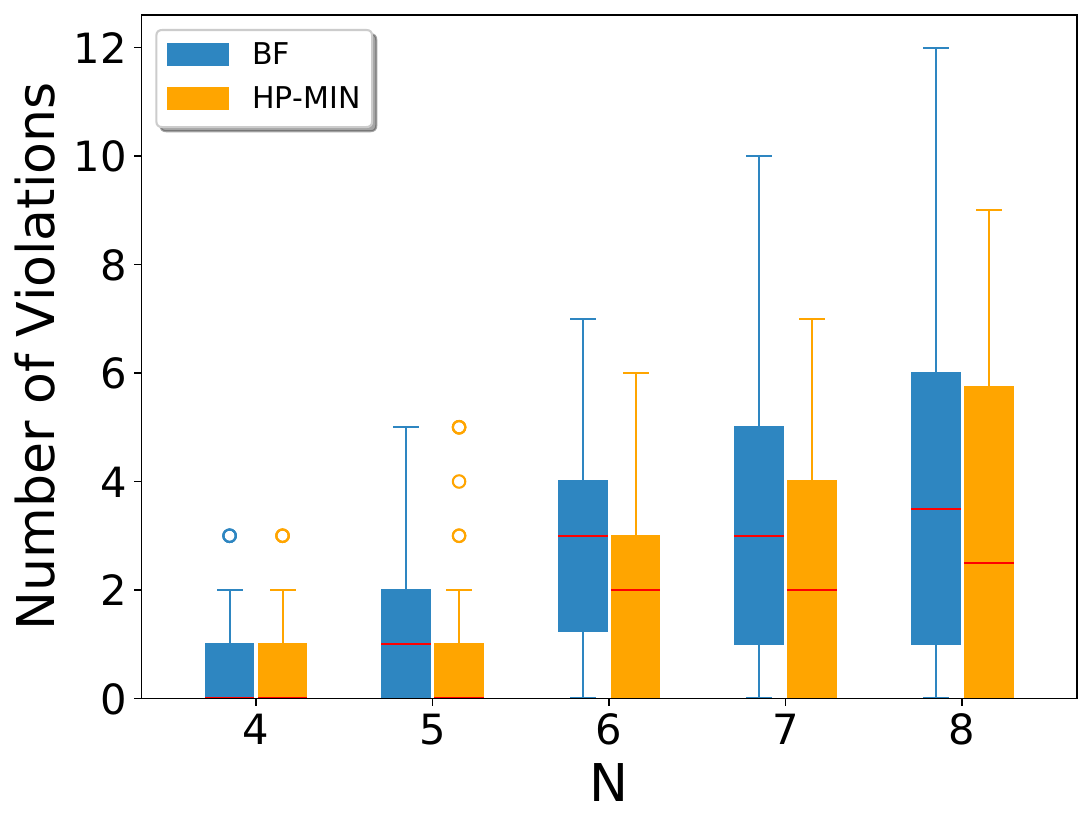}
        \captionsetup{type=figure}
        \caption{Property \ref{prop.ce}: Number of Violations}
        \label{fig : prop1_violations_app}
    \end{subfigure}\hfill\hfill%
    \begin{subfigure}[c]{0.64\textwidth}
        \includegraphics[width=\textwidth, height=0.195\textheight]{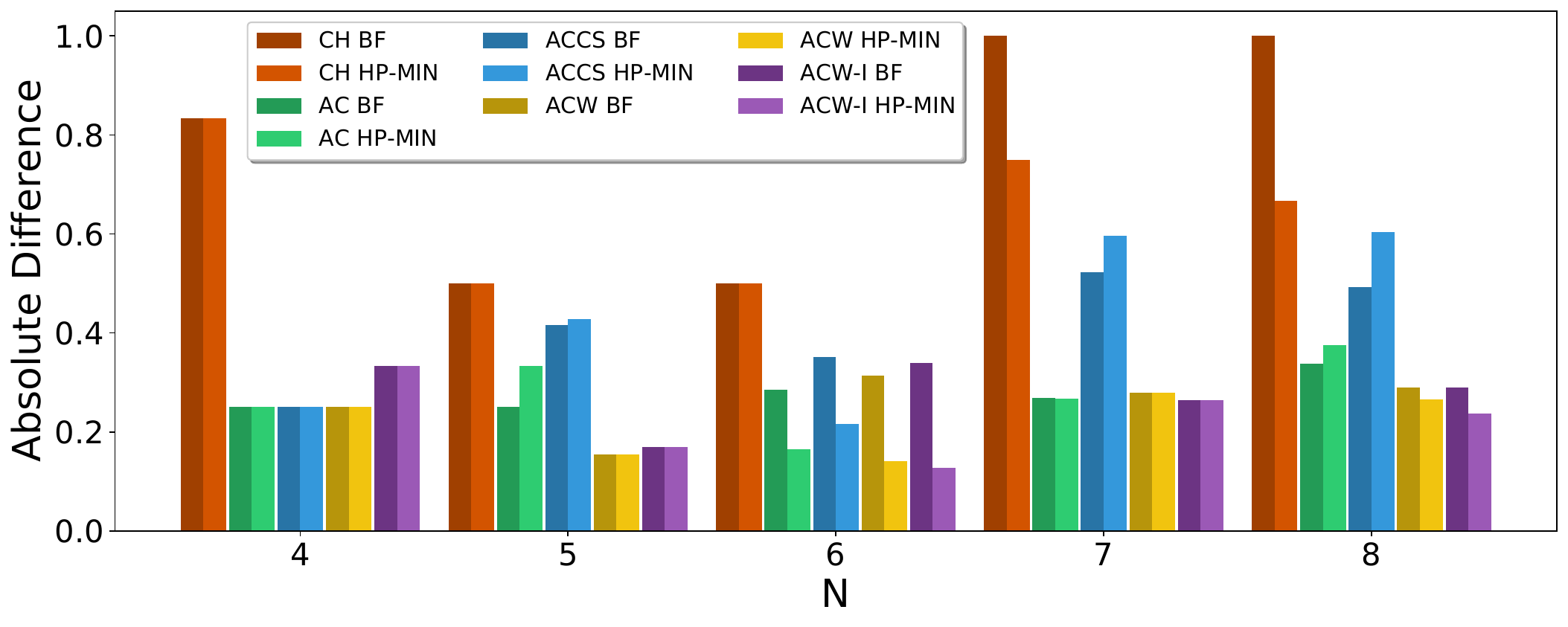}
        \captionsetup{type=figure}
        \caption{Property \ref{prop.ce}: Impact on Responsibility}
        \label{fig : prop1_resp_app}
    \end{subfigure}\\
    \begin{subfigure}[c]{0.32\textwidth}
        \includegraphics[width=\textwidth, height=0.195\textheight]{./figures/causes/prop3.pdf}
        \captionsetup{type=figure}
        \caption{Property \ref{prop.min}: Number of Violations}
        \label{fig : prop3_violations_app}
    \end{subfigure}\hfill%
    \begin{subfigure}[c]{0.32\textwidth}
        \includegraphics[width=\textwidth, height=0.195\textheight]{./figures/causes/num_cont.pdf}
        \captionsetup{type=figure}
        \subcaption{Actual Cause-Contingency Pairs}
        \label{fig : num_ac_cont_app}
    \end{subfigure}\hfill%
    \begin{subfigure}[c]{0.32\textwidth}
        \includegraphics[width=\textwidth, height=0.195\textheight]{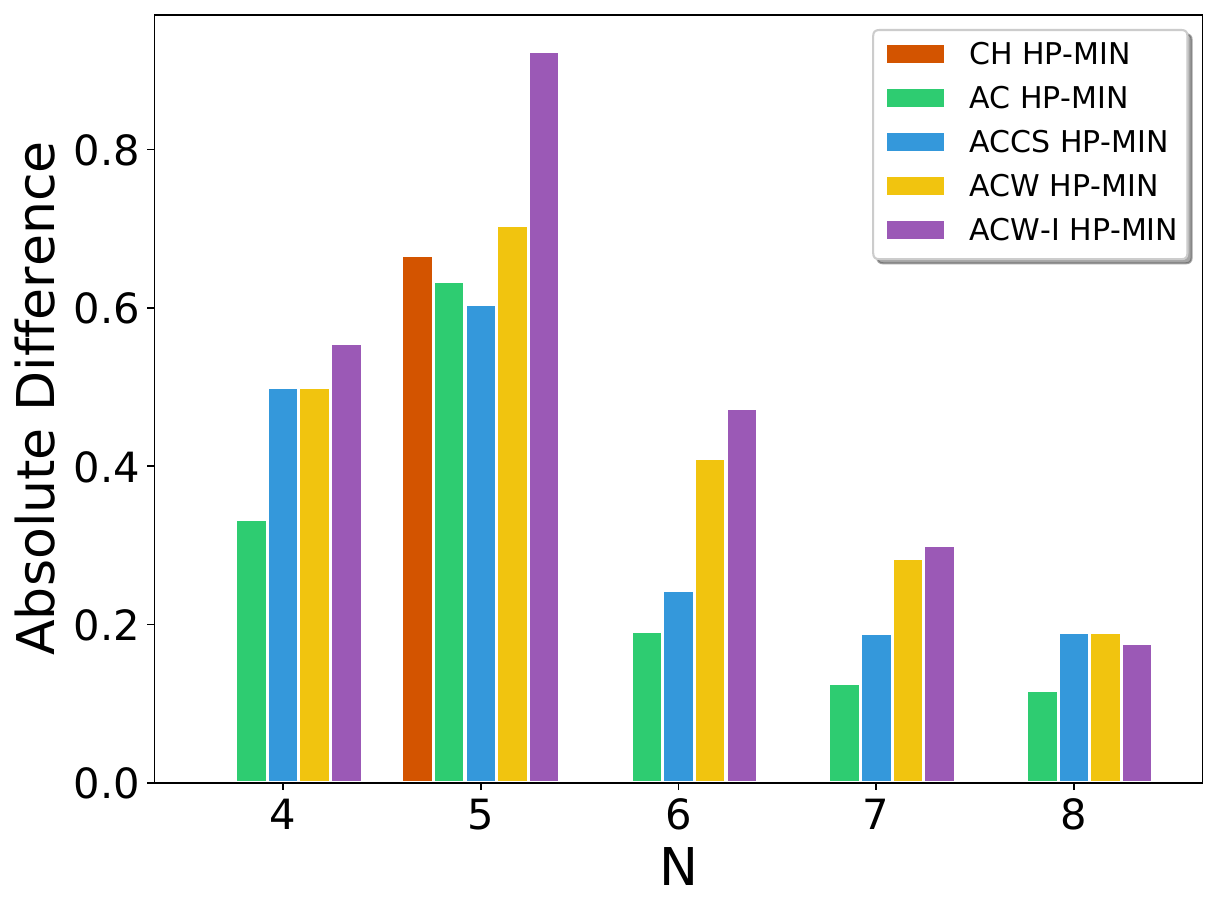}
        \captionsetup{type=figure}
        \caption{Property \ref{prop.min}: Impact on Responsibility}
        \label{fig : prop3_resp_app}
    \end{subfigure}
\Description{Plots \ref{fig : prop1_violations_app} and \ref{fig : prop3_violations_app} show the number of violations of Properties \ref{prop.ce} and \ref{prop.min}. Plots \ref{fig : prop1_resp_app} and \ref{fig : prop3_resp_app} show the impact of these violations on the agents' degrees of responsibility. Plot \ref{fig : num_ac_cont_app} shows the number of distinct actual cause-contingency pairs. Compared to Plots \ref{fig : prop1_violations}, \ref{fig : prop1_resp} and \ref{fig : prop3_resp}, Plots \ref{fig : prop1_violations_app}, \ref{fig : prop1_resp_app} and \ref{fig : prop3_resp_app} have the HP definition replaced by the HP-MIN definitions. All plots are shown over $50$ trajectories per value of $N$.}
\caption{Plots \ref{fig : prop1_violations_app} and \ref{fig : prop3_violations_app} show the number of violations of Properties \ref{prop.ce} and \ref{prop.min}. Plots \ref{fig : prop1_resp_app} and \ref{fig : prop3_resp_app} show the impact of these violations on the agents' degrees of responsibility. Plot \ref{fig : num_ac_cont_app} shows the number of distinct actual cause-contingency pairs. Compared to Plots \ref{fig : prop1_violations}, \ref{fig : prop1_resp} and \ref{fig : prop3_resp}, Plots \ref{fig : prop1_violations_app}, \ref{fig : prop1_resp_app} and \ref{fig : prop3_resp_app} have the HP definition replaced by the HP-MIN definition.}
\end{figure*}

In this section, we present the results from Section \ref{sec.violations} after replacing the HP definition for actual cause with its enhanced version, HP-MIN, which was introduced in Section \ref{sec.prop3_violations}. As expected, Plots \ref{fig : prop1_violations} and \ref{fig : prop1_violations_app} are identical, since the number of Property \ref{prop.ce} violations is not affected by whether the contingency minimality condition is satisfied or not. As a result, the differences in Plots \ref{fig : prop3_violations} and \ref{fig : prop3_violations_app} are insignificant. As mentioned in Section \ref{sec.prop3_violations}, the number of Property \ref{prop.min} violations is considerably reduced when replacing HP with HP-MIN. Although, Plot \ref{fig : prop3_resp_app} (compared to Plot \ref{fig : prop3_resp}) shows a similar tendency for the impact on the agents' degrees of responsibility,\footnote{At least for \textbf{ACCS}, \textbf{ACW} and \textbf{ACW-I} which are the only ones being affected.} it can be seen that this impact is still quite large.

%% file: 7.5_add_ac_comp.tex

\section{Additional Comparison Results of Actual Causality Definitions}\label{sec.add_comp_resuts}

\captionsetup[figure]{belowskip=0pt}
\begin{figure*}
\captionsetup[subfigure]{aboveskip=0pt,belowskip=1pt}
\centering
    \begin{subfigure}{0.24\textwidth}
        \includegraphics[width=\textwidth]{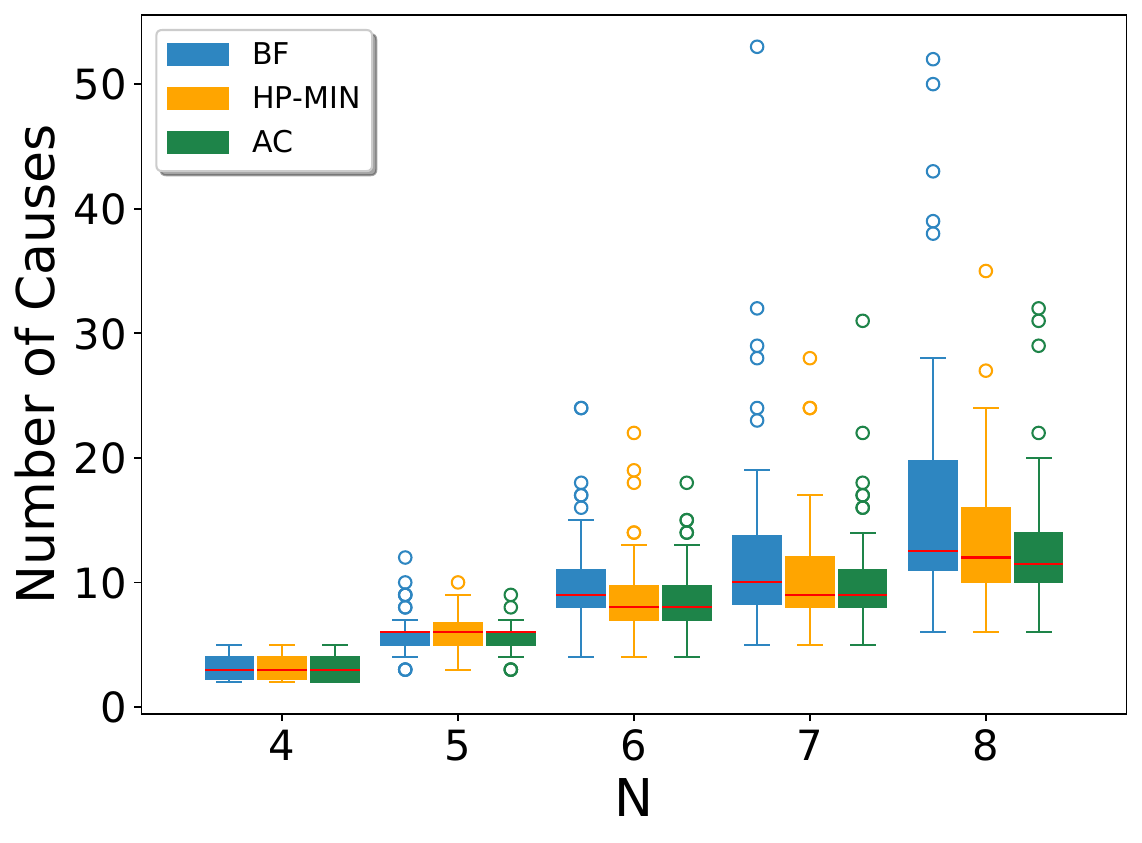}
        \captionsetup{type=figure}
        \caption{Number of Actual Causes}
        \label{fig: num}
    \end{subfigure}
    \begin{subfigure}{0.24\textwidth}
        \includegraphics[width=\textwidth]{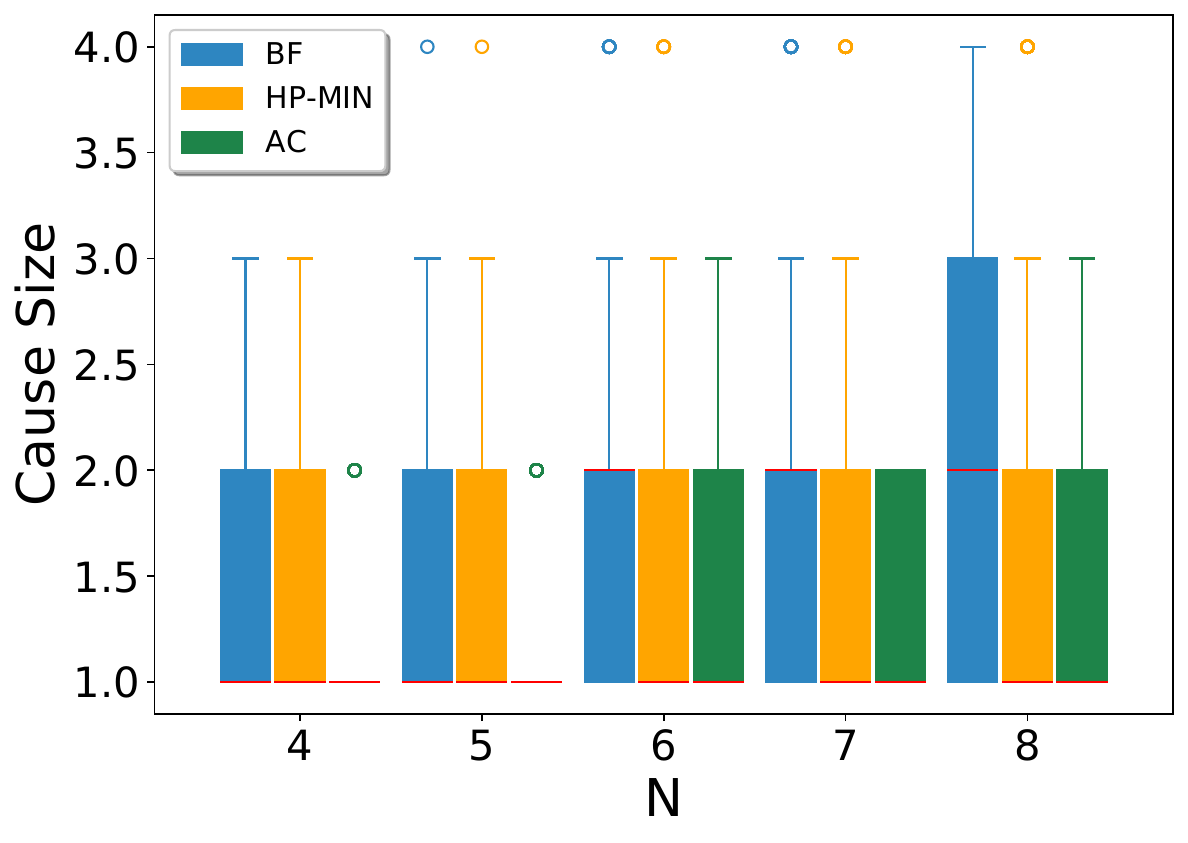}
        \captionsetup{type=figure}
        \caption{Size of Actual Causes}
        \label{fig: size}
    \end{subfigure}
    \begin{subfigure}{0.24\textwidth}
        \includegraphics[width=\textwidth]{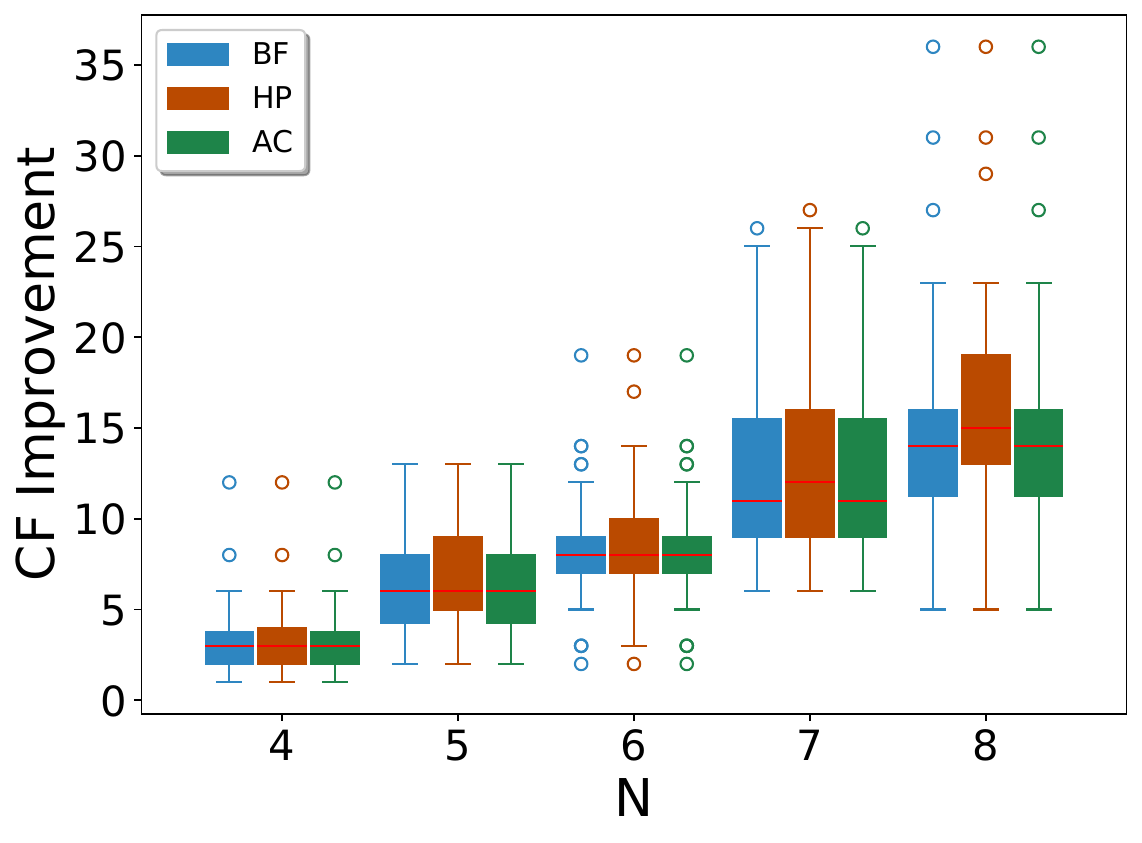}
        \captionsetup{type=figure}
        \caption{CF Improvement}
        \label{fig: cf_impr}
    \end{subfigure}
    \begin{subfigure}{0.24\textwidth}
        \includegraphics[width=\textwidth]{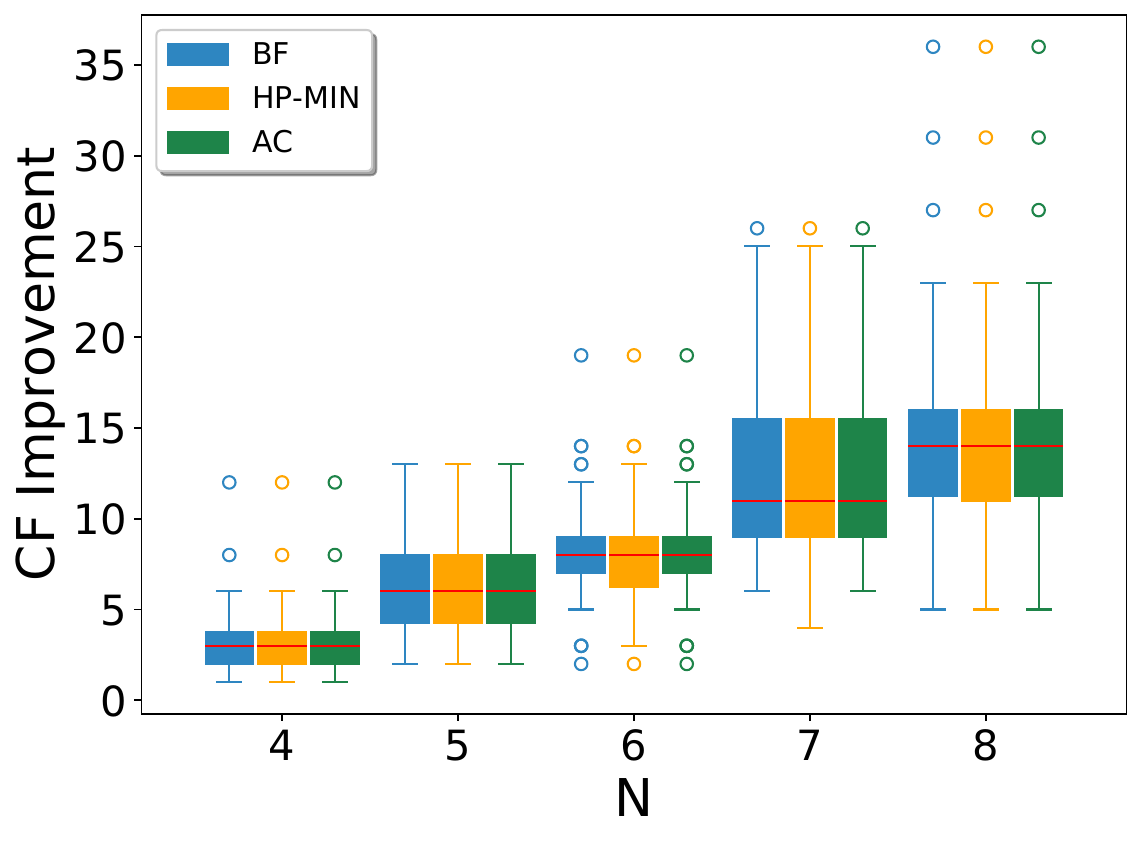}
        \captionsetup{type=figure}
        \subcaption{CF Improvement (HP-MIN)}
        \label{fig: cf_impr_min}
    \end{subfigure}
\Description{Additional Metrics}
\caption{Plots \ref{fig: num} and \ref{fig: size} show the number and size of distinct actual causes. Plots \ref{fig: cf_impr} and \ref{fig: cf_impr_min} show the counterfactual improvement admitted by the actual cause-witness pairs of each actual causality definition.}
\end{figure*}

In this section, we provide some additional empirical insights that we gain by comparing the actual causality definitions from Section \ref{sec.ac_definitions}, in the experimental test-bed of Section \ref{sec.experiments}. 
Plots \ref{fig: num} and \ref{fig: size} display the number of distinct actual causes and their corresponding size (over all sampled trajectories), respectively. As expected, the BF definition admits a larger number of distinct actual causes, and of greater size than the other two definitions, since it is not equipped with the notion of contingencies. 
Plot \ref{fig: cf_impr} shows the counterfactual improvement admitted by the actual cause-witness pairs of each definition. It can be seen that the HP definition provides pairs that admit greater improvement in general. However, the main reason why this is happening, is because HP allows for non-minimal contingencies (see Section \ref{sec.prop3_violations} and Appendix \ref{sec.hp-min}). Plot \ref{fig: cf_impr_min} validates this intuition.

%% file: main.bbl

\begin{thebibliography}{53}


\ifx \showCODEN    \undefined \def \showCODEN     #1{\unskip}     \fi
\ifx \showDOI      \undefined \def \showDOI       #1{#1}\fi
\ifx \showISBNx    \undefined \def \showISBNx     #1{\unskip}     \fi
\ifx \showISBNxiii \undefined \def \showISBNxiii  #1{\unskip}     \fi
\ifx \showISSN     \undefined \def \showISSN      #1{\unskip}     \fi
\ifx \showLCCN     \undefined \def \showLCCN      #1{\unskip}     \fi
\ifx \shownote     \undefined \def \shownote      #1{#1}          \fi
\ifx \showarticletitle \undefined \def \showarticletitle #1{#1}   \fi
\ifx \showURL      \undefined \def \showURL       {\relax}        \fi
\providecommand\bibfield[2]{#2}
\providecommand\bibinfo[2]{#2}
\providecommand\natexlab[1]{#1}
\providecommand\showeprint[2][]{arXiv:#2}

\bibitem[Alechina et~al\mbox{.}(2020)]%
        {alechina2020causality}
\bibfield{author}{\bibinfo{person}{Natasha Alechina},
  \bibinfo{person}{Joseph~Y. Halpern}, {and} \bibinfo{person}{Brian Logan}.}
  \bibinfo{year}{2020}\natexlab{}.
\newblock \showarticletitle{Causality, responsibility and blame in team plans}.
\newblock \bibinfo{journal}{\emph{arXiv preprint arXiv:2005.10297}}
  (\bibinfo{year}{2020}).
\newblock


\bibitem[Baier et~al\mbox{.}(2021a)]%
        {baier2021game}
\bibfield{author}{\bibinfo{person}{Christel Baier}, \bibinfo{person}{Florian
  Funke}, {and} \bibinfo{person}{Rupak Majumdar}.}
  \bibinfo{year}{2021}\natexlab{a}.
\newblock \showarticletitle{A game-theoretic account of responsibility
  allocation}.
\newblock \bibinfo{journal}{\emph{arXiv preprint arXiv:2105.09129}}
  (\bibinfo{year}{2021}).
\newblock


\bibitem[Baier et~al\mbox{.}(2021b)]%
        {baier2021responsibility}
\bibfield{author}{\bibinfo{person}{Christel Baier}, \bibinfo{person}{Florian
  Funke}, {and} \bibinfo{person}{Rupak Majumdar}.}
  \bibinfo{year}{2021}\natexlab{b}.
\newblock \showarticletitle{Responsibility attribution in parameterized
  Markovian models}. In \bibinfo{booktitle}{\emph{Proc. of the 35th AAAI
  Conference on Artificial Intelligence (AAAI)}}.
  \bibinfo{pages}{11734--11743}.
\newblock


\bibitem[Banzhaf~III(1964)]%
        {banzhaf1964weighted}
\bibfield{author}{\bibinfo{person}{John~F. Banzhaf~III}.}
  \bibinfo{year}{1964}\natexlab{}.
\newblock \showarticletitle{Weighted voting doesn't work: {A} mathematical
  analysis}.
\newblock \bibinfo{journal}{\emph{Rutgers Law Review}}  \bibinfo{volume}{19}
  (\bibinfo{year}{1964}), \bibinfo{pages}{317}.
\newblock


\bibitem[Banzhaf~III(1968)]%
        {banzhaf1968one}
\bibfield{author}{\bibinfo{person}{John~F. Banzhaf~III}.}
  \bibinfo{year}{1968}\natexlab{}.
\newblock \showarticletitle{One man, 3.312 votes: a mathematical analysis of
  the Electoral College}.
\newblock \bibinfo{journal}{\emph{Villanova Law Review}}  \bibinfo{volume}{13}
  (\bibinfo{year}{1968}), \bibinfo{pages}{304}.
\newblock


\bibitem[Bernstein et~al\mbox{.}(2002)]%
        {bernstein2002complexity}
\bibfield{author}{\bibinfo{person}{Daniel~S. Bernstein},
  \bibinfo{person}{Robert Givan}, \bibinfo{person}{Neil Immerman}, {and}
  \bibinfo{person}{Shlomo Zilberstein}.} \bibinfo{year}{2002}\natexlab{}.
\newblock \showarticletitle{The complexity of decentralized control of Markov
  decision processes}.
\newblock \bibinfo{journal}{\emph{Mathematics of operations research}}
  \bibinfo{volume}{27}, \bibinfo{number}{4} (\bibinfo{year}{2002}),
  \bibinfo{pages}{819--840}.
\newblock


\bibitem[Buesing et~al\mbox{.}(2018)]%
        {buesing2018woulda}
\bibfield{author}{\bibinfo{person}{Lars Buesing}, \bibinfo{person}{Theophane
  Weber}, \bibinfo{person}{Yori Zwols}, \bibinfo{person}{Sebastien Racaniere},
  \bibinfo{person}{Arthur Guez}, \bibinfo{person}{Jean-Baptiste Lespiau}, {and}
  \bibinfo{person}{Nicolas Heess}.} \bibinfo{year}{2018}\natexlab{}.
\newblock \showarticletitle{Woulda, coulda, shoulda: Counterfactually-guided
  policy search}.
\newblock \bibinfo{journal}{\emph{arXiv preprint arXiv:1811.06272}}
  (\bibinfo{year}{2018}).
\newblock


\bibitem[Chockler and Halpern(2004)]%
        {chockler2004responsibility}
\bibfield{author}{\bibinfo{person}{Hana Chockler} {and}
  \bibinfo{person}{Joseph~Y. Halpern}.} \bibinfo{year}{2004}\natexlab{}.
\newblock \showarticletitle{Responsibility and blame: A structural-model
  approach}.
\newblock \bibinfo{journal}{\emph{Journal of Artificial Intelligence Research}}
   \bibinfo{volume}{22} (\bibinfo{year}{2004}), \bibinfo{pages}{93--115}.
\newblock


\bibitem[Coeckelbergh(2020)]%
        {coeckelbergh2020artificial}
\bibfield{author}{\bibinfo{person}{Mark Coeckelbergh}.}
  \bibinfo{year}{2020}\natexlab{}.
\newblock \showarticletitle{Artificial intelligence, responsibility
  attribution, and a relational justification of explainability}.
\newblock \bibinfo{journal}{\emph{Science and engineering ethics}}
  \bibinfo{volume}{26}, \bibinfo{number}{4} (\bibinfo{year}{2020}),
  \bibinfo{pages}{2051--2068}.
\newblock


\bibitem[Datta et~al\mbox{.}(2015)]%
        {datta2015program}
\bibfield{author}{\bibinfo{person}{Anupam Datta}, \bibinfo{person}{Deepak
  Garg}, \bibinfo{person}{Dilsun Kaynar}, \bibinfo{person}{Divya Sharma}, {and}
  \bibinfo{person}{Arunesh Sinha}.} \bibinfo{year}{2015}\natexlab{}.
\newblock \showarticletitle{Program actions as actual causes: A building block
  for accountability}. In \bibinfo{booktitle}{\emph{2015 IEEE 28th Computer
  Security Foundations Symposium}}. \bibinfo{pages}{261--275}.
\newblock


\bibitem[{European Commission}(2019)]%
        {ethicsEU}
\bibfield{author}{\bibinfo{person}{{European Commission}}.}
  \bibinfo{year}{2019}\natexlab{}.
\newblock \bibinfo{title}{{Ethics Guidelines for Trustworthy Artificial
  Intelligence}}.
\newblock \bibinfo{howpublished}{URL:
  \url{https://ec.europa.eu/digital-single-market/en/news/ethics-guidelines-trustworthy-ai}}.
\newblock
\newblock
\shownote{[Online; accessed 15-January-2021]}.


\bibitem[Friedenberg and Halpern(2019)]%
        {friedenberg2019blameworthiness}
\bibfield{author}{\bibinfo{person}{Meir Friedenberg} {and}
  \bibinfo{person}{Joseph~Y. Halpern}.} \bibinfo{year}{2019}\natexlab{}.
\newblock \showarticletitle{Blameworthiness in multi-agent settings}. In
  \bibinfo{booktitle}{\emph{Proceedings of the AAAI Conference on Artificial
  Intelligence}}, Vol.~\bibinfo{volume}{33}. \bibinfo{pages}{525--532}.
\newblock


\bibitem[Grimes and Dror(2013)]%
        {grimes2013observations}
\bibfield{author}{\bibinfo{person}{Mark Grimes} {and} \bibinfo{person}{Moshe
  Dror}.} \bibinfo{year}{2013}\natexlab{}.
\newblock \showarticletitle{Observations on strategies for Goofspiel}. In
  \bibinfo{booktitle}{\emph{2013 IEEE Conference on Computational Inteligence
  in Games (CIG)}}. \bibinfo{pages}{1--2}.
\newblock


\bibitem[Hall(2007)]%
        {hall2007structural}
\bibfield{author}{\bibinfo{person}{Ned Hall}.} \bibinfo{year}{2007}\natexlab{}.
\newblock \showarticletitle{Structural equations and causation}.
\newblock \bibinfo{journal}{\emph{Philosophical Studies}}
  \bibinfo{volume}{132}, \bibinfo{number}{1} (\bibinfo{year}{2007}),
  \bibinfo{pages}{109--136}.
\newblock


\bibitem[Halpern(2008)]%
        {halpern2008defaults}
\bibfield{author}{\bibinfo{person}{Joseph~Y. Halpern}.}
  \bibinfo{year}{2008}\natexlab{}.
\newblock \showarticletitle{Defaults and normality in causal structures}. In
  \bibinfo{booktitle}{\emph{Eleventh International Conference of Knowledge
  Representation and Reasoning}}. \bibinfo{pages}{198--208}.
\newblock


\bibitem[Halpern(2015)]%
        {halpern2015modification}
\bibfield{author}{\bibinfo{person}{Joseph~Y. Halpern}.}
  \bibinfo{year}{2015}\natexlab{}.
\newblock \showarticletitle{A modification of the Halpern-Pearl definition of
  causality}. In \bibinfo{booktitle}{\emph{Twenty-Fourth International Joint
  Conference on Artificial Intelligence}}. \bibinfo{pages}{3022–3033}.
\newblock


\bibitem[Halpern(2016)]%
        {halpern2016actual}
\bibfield{author}{\bibinfo{person}{Joseph~Y. Halpern}.}
  \bibinfo{year}{2016}\natexlab{}.
\newblock \bibinfo{booktitle}{\emph{Actual causality}}.
\newblock \bibinfo{publisher}{MIT Press}.
\newblock


\bibitem[Halpern and Hitchcock(2015)]%
        {halpern2015graded}
\bibfield{author}{\bibinfo{person}{Joseph~Y. Halpern} {and}
  \bibinfo{person}{Christopher Hitchcock}.} \bibinfo{year}{2015}\natexlab{}.
\newblock \showarticletitle{Graded causation and defaults}.
\newblock \bibinfo{journal}{\emph{The British Journal for the Philosophy of
  Science}} \bibinfo{volume}{66}, \bibinfo{number}{2} (\bibinfo{year}{2015}),
  \bibinfo{pages}{413--457}.
\newblock


\bibitem[Halpern and Kleiman-Weiner(2018)]%
        {halpern2018towards}
\bibfield{author}{\bibinfo{person}{Joseph~Y. Halpern} {and}
  \bibinfo{person}{Max Kleiman-Weiner}.} \bibinfo{year}{2018}\natexlab{}.
\newblock \showarticletitle{Towards formal definitions of blameworthiness,
  intention, and moral responsibility}. In
  \bibinfo{booktitle}{\emph{Proceedings of the AAAI Conference on Artificial
  Intelligence}}, Vol.~\bibinfo{volume}{32}.
\newblock


\bibitem[Halpern and Pearl(2001)]%
        {halpern2001causes}
\bibfield{author}{\bibinfo{person}{Joseph~Y. Halpern} {and}
  \bibinfo{person}{Judea Pearl}.} \bibinfo{year}{2001}\natexlab{}.
\newblock \showarticletitle{Causes and explanations: A structural-model
  approach. Part I: Causes.}. In \bibinfo{booktitle}{\emph{Proceedings of the
  Seventeenth Conference on Uncertainty in Artificial Intelligence}}.
  \bibinfo{pages}{194--202}.
\newblock


\bibitem[Halpern and Pearl(2005)]%
        {halpern2005causes}
\bibfield{author}{\bibinfo{person}{Joseph~Y. Halpern} {and}
  \bibinfo{person}{Judea Pearl}.} \bibinfo{year}{2005}\natexlab{}.
\newblock \showarticletitle{Causes and explanations: A structural-model
  approach. Part I: Causes.}
\newblock \bibinfo{journal}{\emph{British Journal for the Philosophy of
  Science}} \bibinfo{volume}{56}, \bibinfo{number}{4} (\bibinfo{year}{2005}).
\newblock


\bibitem[Hart and Honor{\'e}(1985)]%
        {hart1985causation}
\bibfield{author}{\bibinfo{person}{Herbert Lionel~Adolphus Hart} {and}
  \bibinfo{person}{Tony Honor{\'e}}.} \bibinfo{year}{1985}\natexlab{}.
\newblock \bibinfo{booktitle}{\emph{Causation in the Law}}.
\newblock \bibinfo{publisher}{OUP Oxford}.
\newblock


\bibitem[Harutyunyan et~al\mbox{.}(2019)]%
        {harutyunyan2019hindsight}
\bibfield{author}{\bibinfo{person}{Anna Harutyunyan}, \bibinfo{person}{Will
  Dabney}, \bibinfo{person}{Thomas Mesnard}, \bibinfo{person}{Mohammad
  Gheshlaghi~Azar}, \bibinfo{person}{Bilal Piot}, \bibinfo{person}{Nicolas
  Heess}, \bibinfo{person}{Hado~P. van Hasselt}, \bibinfo{person}{Gregory
  Wayne}, \bibinfo{person}{Satinder Singh}, \bibinfo{person}{Doina Precup},
  {and} \bibinfo{person}{et.al.}} \bibinfo{year}{2019}\natexlab{}.
\newblock \showarticletitle{Hindsight credit assignment}.
\newblock \bibinfo{journal}{\emph{Advances in Neural Information Processing
  Systems}}  \bibinfo{volume}{32} (\bibinfo{year}{2019}),
  \bibinfo{pages}{12467–12476}.
\newblock


\bibitem[Hennes et~al\mbox{.}(2020)]%
        {hennes2020neural}
\bibfield{author}{\bibinfo{person}{Daniel Hennes}, \bibinfo{person}{Dustin
  Morrill}, \bibinfo{person}{Shayegan Omidshafiei}, \bibinfo{person}{R{\'e}mi
  Munos}, \bibinfo{person}{Julien Perolat}, \bibinfo{person}{Marc Lanctot},
  \bibinfo{person}{Audrunas Gruslys}, \bibinfo{person}{Jean-Baptiste Lespiau},
  \bibinfo{person}{Paavo Parmas}, \bibinfo{person}{Edgar
  Du{\'e}{\~n}ez-Guzm{\'a}n}, {et~al\mbox{.}}} \bibinfo{year}{2020}\natexlab{}.
\newblock \showarticletitle{Neural replicator dynamics: Multiagent learning via
  hedging policy gradients}. In \bibinfo{booktitle}{\emph{Proceedings of the
  19th International Conference on Autonomous Agents and MultiAgent Systems}}.
  \bibinfo{pages}{492--501}.
\newblock


\bibitem[Hiddleston(2005)]%
        {hiddleston2005causal}
\bibfield{author}{\bibinfo{person}{Eric Hiddleston}.}
  \bibinfo{year}{2005}\natexlab{}.
\newblock \showarticletitle{Causal powers}.
\newblock \bibinfo{journal}{\emph{The British journal for the philosophy of
  science}} \bibinfo{volume}{56}, \bibinfo{number}{1} (\bibinfo{year}{2005}),
  \bibinfo{pages}{27--59}.
\newblock


\bibitem[Hitchcock(2001)]%
        {hitchcock2001intransitivity}
\bibfield{author}{\bibinfo{person}{Christopher Hitchcock}.}
  \bibinfo{year}{2001}\natexlab{}.
\newblock \showarticletitle{The intransitivity of causation revealed in
  equations and graphs}.
\newblock \bibinfo{journal}{\emph{The Journal of Philosophy}}
  \bibinfo{volume}{98}, \bibinfo{number}{6} (\bibinfo{year}{2001}),
  \bibinfo{pages}{273--299}.
\newblock


\bibitem[Hitchcock(2007)]%
        {hitchcock2007prevention}
\bibfield{author}{\bibinfo{person}{Christopher Hitchcock}.}
  \bibinfo{year}{2007}\natexlab{}.
\newblock \showarticletitle{Prevention, preemption, and the principle of
  sufficient reason}.
\newblock \bibinfo{journal}{\emph{The Philosophical Review}}
  \bibinfo{volume}{116}, \bibinfo{number}{4} (\bibinfo{year}{2007}),
  \bibinfo{pages}{495--532}.
\newblock


\bibitem[Hume(2000)]%
        {hume2000enquiry}
\bibfield{author}{\bibinfo{person}{David Hume}.}
  \bibinfo{year}{2000}\natexlab{}.
\newblock \bibinfo{booktitle}{\emph{An enquiry concerning human understanding:
  A critical edition}}.
\newblock \bibinfo{publisher}{Oxford University Press}.
\newblock


\bibitem[Ibrahim(2021)]%
        {ibrahim2021actual}
\bibfield{author}{\bibinfo{person}{Amjad Ibrahim}.}
  \bibinfo{year}{2021}\natexlab{}.
\newblock \emph{\bibinfo{title}{An actual causality framework for accountable
  systems}}.
\newblock \bibinfo{thesistype}{Ph.\,D. Dissertation}.
  \bibinfo{school}{Technische Universit{\"a}t M{\"u}nchen}.
\newblock


\bibitem[Jain and Mahdian(2007)]%
        {jain2007cost}
\bibfield{author}{\bibinfo{person}{Kamal Jain} {and} \bibinfo{person}{Mohammad
  Mahdian}.} \bibinfo{year}{2007}\natexlab{}.
\newblock \showarticletitle{Cost sharing}.
\newblock \bibinfo{journal}{\emph{Algorithmic Game Theory}}
  \bibinfo{volume}{15} (\bibinfo{year}{2007}), \bibinfo{pages}{385--410}.
\newblock


\bibitem[Lanctot et~al\mbox{.}(2019)]%
        {lanctot2019openspiel}
\bibfield{author}{\bibinfo{person}{Marc Lanctot}, \bibinfo{person}{Edward
  Lockhart}, \bibinfo{person}{Jean-Baptiste Lespiau}, \bibinfo{person}{Vinicius
  Zambaldi}, \bibinfo{person}{Satyaki Upadhyay}, \bibinfo{person}{Julien
  P{\'e}rolat}, \bibinfo{person}{Sriram Srinivasan}, \bibinfo{person}{Finbarr
  Timbers}, \bibinfo{person}{Karl Tuyls}, \bibinfo{person}{Shayegan
  Omidshafiei}, {et~al\mbox{.}}} \bibinfo{year}{2019}\natexlab{}.
\newblock \showarticletitle{OpenSpiel: A framework for reinforcement learning
  in games}.
\newblock \bibinfo{journal}{\emph{arXiv preprint arXiv:1908.09453}}
  (\bibinfo{year}{2019}).
\newblock


\bibitem[Lewis(1974)]%
        {lewis1974causation}
\bibfield{author}{\bibinfo{person}{David Lewis}.}
  \bibinfo{year}{1974}\natexlab{}.
\newblock \showarticletitle{Causation}.
\newblock \bibinfo{journal}{\emph{The Journal of Philosophy}}
  \bibinfo{volume}{70}, \bibinfo{number}{17} (\bibinfo{year}{1974}),
  \bibinfo{pages}{556--567}.
\newblock


\bibitem[Lewis(2013)]%
        {lewis2013counterfactuals}
\bibfield{author}{\bibinfo{person}{David Lewis}.}
  \bibinfo{year}{2013}\natexlab{}.
\newblock \bibinfo{booktitle}{\emph{Counterfactuals}}.
\newblock \bibinfo{publisher}{John Wiley \& Sons}.
\newblock


\bibitem[Madumal et~al\mbox{.}(2020)]%
        {madumal2020explainable}
\bibfield{author}{\bibinfo{person}{Prashan Madumal}, \bibinfo{person}{Tim
  Miller}, \bibinfo{person}{Liz Sonenberg}, {and} \bibinfo{person}{Frank
  Vetere}.} \bibinfo{year}{2020}\natexlab{}.
\newblock \showarticletitle{Explainable reinforcement learning through a causal
  lens}. In \bibinfo{booktitle}{\emph{Proceedings of the AAAI conference on
  artificial intelligence}}. \bibinfo{pages}{2493--2500}.
\newblock


\bibitem[Mesnard et~al\mbox{.}(2021)]%
        {mesnard2021counterfactual}
\bibfield{author}{\bibinfo{person}{Thomas Mesnard}, \bibinfo{person}{Theophane
  Weber}, \bibinfo{person}{Fabio Viola}, \bibinfo{person}{Shantanu Thakoor},
  \bibinfo{person}{Alaa Saade}, \bibinfo{person}{Anna Harutyunyan},
  \bibinfo{person}{Will Dabney}, \bibinfo{person}{Thomas~S Stepleton},
  \bibinfo{person}{Nicolas Heess}, \bibinfo{person}{Arthur Guez}, {and}
  \bibinfo{person}{et.al.}} \bibinfo{year}{2021}\natexlab{}.
\newblock \showarticletitle{Counterfactual credit assignment in model-free
  reinforcement learning}. In \bibinfo{booktitle}{\emph{International
  Conference on Machine Learning}}. \bibinfo{pages}{7654--7664}.
\newblock


\bibitem[Micalizio et~al\mbox{.}(2004)]%
        {micalizio2004line}
\bibfield{author}{\bibinfo{person}{Roberto Micalizio}, \bibinfo{person}{Pietro
  Torasso}, {and} \bibinfo{person}{Gianluca Torta}.}
  \bibinfo{year}{2004}\natexlab{}.
\newblock \showarticletitle{On-line monitoring and diagnosis of multi-agent
  systems: A model based approach}. In \bibinfo{booktitle}{\emph{Proceedings of
  the 16th Eureopean Conference on Artificial Intelligence}},
  Vol.~\bibinfo{volume}{16}. \bibinfo{pages}{848}.
\newblock


\bibitem[Oberst and Sontag(2019)]%
        {oberst2019counterfactual}
\bibfield{author}{\bibinfo{person}{Michael Oberst} {and} \bibinfo{person}{David
  Sontag}.} \bibinfo{year}{2019}\natexlab{}.
\newblock \showarticletitle{Counterfactual off-policy evaluation with
  gumbel-max structural causal models}. In
  \bibinfo{booktitle}{\emph{International Conference on Machine Learning}}.
  \bibinfo{pages}{4881--4890}.
\newblock


\bibitem[Oliehoek and Amato(2016)]%
        {oliehoek2016concise}
\bibfield{author}{\bibinfo{person}{Frans~A. Oliehoek} {and}
  \bibinfo{person}{Christopher Amato}.} \bibinfo{year}{2016}\natexlab{}.
\newblock \bibinfo{booktitle}{\emph{A concise introduction to decentralized
  POMDPs}}.
\newblock \bibinfo{publisher}{Springer}.
\newblock


\bibitem[Pearl(1995)]%
        {pearl1995causal}
\bibfield{author}{\bibinfo{person}{Judea Pearl}.}
  \bibinfo{year}{1995}\natexlab{}.
\newblock \showarticletitle{Causal diagrams for empirical research}.
\newblock \bibinfo{journal}{\emph{Biometrika}} \bibinfo{volume}{82},
  \bibinfo{number}{4} (\bibinfo{year}{1995}), \bibinfo{pages}{669--688}.
\newblock


\bibitem[Pearl(1998)]%
        {pearl1998definition}
\bibfield{author}{\bibinfo{person}{Judea Pearl}.}
  \bibinfo{year}{1998}\natexlab{}.
\newblock \bibinfo{booktitle}{\emph{On the definition of actual cause}}.
\newblock \bibinfo{type}{Technical Report R-259}.
  \bibinfo{institution}{Department of Computer Science, University of
  California, Los Angeles}.
\newblock


\bibitem[Pearl(2009)]%
        {pearl2009causality}
\bibfield{author}{\bibinfo{person}{Judea Pearl}.}
  \bibinfo{year}{2009}\natexlab{}.
\newblock \bibinfo{booktitle}{\emph{Causality}}.
\newblock \bibinfo{publisher}{Cambridge University Press}.
\newblock


\bibitem[Peters et~al\mbox{.}(2017)]%
        {peters2017elements}
\bibfield{author}{\bibinfo{person}{Jonas Peters}, \bibinfo{person}{Dominik
  Janzing}, {and} \bibinfo{person}{Bernhard Sch{\"o}lkopf}.}
  \bibinfo{year}{2017}\natexlab{}.
\newblock \bibinfo{booktitle}{\emph{Elements of causal inference: Foundations
  and learning algorithms}}.
\newblock \bibinfo{publisher}{The MIT Press}.
\newblock


\bibitem[Poel(2011)]%
        {poel2011relation}
\bibfield{author}{\bibinfo{person}{Ibo van~de Poel}.}
  \bibinfo{year}{2011}\natexlab{}.
\newblock \showarticletitle{The relation between forward-looking and
  backward-looking responsibility}.
\newblock In \bibinfo{booktitle}{\emph{Moral responsibility}}.
  \bibinfo{publisher}{Springer}, \bibinfo{pages}{37--52}.
\newblock


\bibitem[Rhoads and Bartholdi(2012)]%
        {rhoads2012computer}
\bibfield{author}{\bibinfo{person}{Glenn~C. Rhoads} {and}
  \bibinfo{person}{Laurent Bartholdi}.} \bibinfo{year}{2012}\natexlab{}.
\newblock \showarticletitle{Computer solution to the game of pure strategy}.
\newblock \bibinfo{journal}{\emph{Games}} \bibinfo{volume}{3},
  \bibinfo{number}{4} (\bibinfo{year}{2012}), \bibinfo{pages}{150--156}.
\newblock


\bibitem[Ross(1971)]%
        {ross1971goofspiel}
\bibfield{author}{\bibinfo{person}{Sheldon~M. Ross}.}
  \bibinfo{year}{1971}\natexlab{}.
\newblock \showarticletitle{Goofspiel—the game of pure strategy}.
\newblock \bibinfo{journal}{\emph{Journal of Applied Probability}}
  \bibinfo{volume}{8}, \bibinfo{number}{3} (\bibinfo{year}{1971}),
  \bibinfo{pages}{621--625}.
\newblock


\bibitem[Shapley(2016)]%
        {shapley201617}
\bibfield{author}{\bibinfo{person}{Lloyd~S. Shapley}.}
  \bibinfo{year}{2016}\natexlab{}.
\newblock \bibinfo{booktitle}{\emph{17. A value for n-person games}}.
\newblock \bibinfo{publisher}{Princeton University Press}.
\newblock


\bibitem[Shapley and Shubik(1954)]%
        {shapley1954method}
\bibfield{author}{\bibinfo{person}{Lloyd~S. Shapley} {and}
  \bibinfo{person}{Martin Shubik}.} \bibinfo{year}{1954}\natexlab{}.
\newblock \showarticletitle{A method for evaluating the distribution of power
  in a committee system}.
\newblock \bibinfo{journal}{\emph{The American Political Science Review}}
  \bibinfo{volume}{48}, \bibinfo{number}{3} (\bibinfo{year}{1954}),
  \bibinfo{pages}{787--792}.
\newblock


\bibitem[Triantafyllou et~al\mbox{.}(2021)]%
        {triantafyllou2021blame}
\bibfield{author}{\bibinfo{person}{Stelios Triantafyllou},
  \bibinfo{person}{Adish Singla}, {and} \bibinfo{person}{Goran Radanovic}.}
  \bibinfo{year}{2021}\natexlab{}.
\newblock \showarticletitle{On blame attribution for accountable multi-agent
  sequential decision making}.
\newblock \bibinfo{journal}{\emph{Advances in Neural Information Processing
  Systems}}  \bibinfo{volume}{34} (\bibinfo{year}{2021}).
\newblock


\bibitem[Tsirtsis et~al\mbox{.}(2021)]%
        {tsirtsis2021counterfactual}
\bibfield{author}{\bibinfo{person}{Stratis Tsirtsis}, \bibinfo{person}{Abir
  De}, {and} \bibinfo{person}{Manuel Rodriguez}.}
  \bibinfo{year}{2021}\natexlab{}.
\newblock \showarticletitle{Counterfactual explanations in sequential decision
  making under uncertainty}.
\newblock \bibinfo{journal}{\emph{Advances in Neural Information Processing
  Systems}}  \bibinfo{volume}{34} (\bibinfo{year}{2021}).
\newblock


\bibitem[Von~Neumann and Morgenstern(2007)]%
        {von2007theory}
\bibfield{author}{\bibinfo{person}{John Von~Neumann} {and}
  \bibinfo{person}{Oskar Morgenstern}.} \bibinfo{year}{2007}\natexlab{}.
\newblock \bibinfo{booktitle}{\emph{Theory of games and economic behavior
  (commemorative edition)}}.
\newblock \bibinfo{publisher}{Princeton University Press}.
\newblock


\bibitem[Witteveen et~al\mbox{.}(2005)]%
        {witteveen2005diagnosis}
\bibfield{author}{\bibinfo{person}{Cees Witteveen}, \bibinfo{person}{Nico
  Roos}, \bibinfo{person}{Roman van~der Krogt}, {and} \bibinfo{person}{Mathijs
  de Weerdt}.} \bibinfo{year}{2005}\natexlab{}.
\newblock \showarticletitle{Diagnosis of single and multi-agent plans}. In
  \bibinfo{booktitle}{\emph{Fourth International Joint Conference on Autonomous
  Agents and Multiagent Systems}}. \bibinfo{pages}{805--812}.
\newblock


\bibitem[Wright(1985)]%
        {wright1985causation}
\bibfield{author}{\bibinfo{person}{Richard~W. Wright}.}
  \bibinfo{year}{1985}\natexlab{}.
\newblock \showarticletitle{Causation in tort law}.
\newblock \bibinfo{journal}{\emph{Calif. L. Rev.}}  \bibinfo{volume}{73}
  (\bibinfo{year}{1985}), \bibinfo{pages}{1735}.
\newblock


\bibitem[Yazdanpanah et~al\mbox{.}(2019)]%
        {yazdanpanah2019strategic}
\bibfield{author}{\bibinfo{person}{Vahid Yazdanpanah}, \bibinfo{person}{Mehdi
  Dastani}, \bibinfo{person}{Natasha Alechina}, \bibinfo{person}{Brian Logan},
  {and} \bibinfo{person}{Wojciech Jamroga}.} \bibinfo{year}{2019}\natexlab{}.
\newblock \showarticletitle{Strategic responsibility under imperfect
  information}. In \bibinfo{booktitle}{\emph{Proceedings of the 18th
  International Conference on Autonomous Agents and Multiagent Systems AAMAS
  2019}}. \bibinfo{pages}{592--600}.
\newblock


\end{thebibliography}
